\documentclass{article}

\usepackage{iclr2027_conference,times}

\usepackage{amsmath,amsfonts,bm}

\def\eqref#1{equation~\ref{#1}}

\def\1{\bm{1}}

\DeclareMathAlphabet{\mathsfit}{\encodingdefault}{\sfdefault}{m}{sl}
\SetMathAlphabet{\mathsfit}{bold}{\encodingdefault}{\sfdefault}{bx}{n}

\usepackage{amsmath}
\usepackage{amssymb}
\usepackage{float}
\usepackage{graphicx}
\usepackage{wrapfig}
\usepackage{booktabs}
\usepackage{multirow}
\usepackage{xcolor}
\usepackage{colortbl}
\definecolor{tabhl}{HTML}{EAF0F8}   

\usepackage{flafter}
\usepackage[section]{placeins}
\usepackage[most]{tcolorbox}
\usepackage{url}
\definecolor{promptrule}{HTML}{D9DEE3}
\definecolor{promptback}{HTML}{F7F8FA}
\newtcolorbox{promptbox}[1][]{%
  enhanced, breakable, colback=promptback, colframe=promptrule,
  boxrule=0.5pt, arc=1.5pt, left=6pt, right=6pt, top=5pt, bottom=5pt,
  fonttitle=\footnotesize\bfseries, coltitle=black,
  attach boxed title to top left={xshift=6pt, yshift=-2pt},
  boxed title style={colback=promptback, colframe=promptrule, boxrule=0.5pt, arc=1.5pt},
  #1}
\usepackage{algorithm}
\usepackage{algorithmic}
\usepackage{natbib}
\usepackage{enumitem}
\usepackage{needspace}
\graphicspath{{./}{figures/}}

\iclrfinalcopy

\begin{document}

\title{TAEC: Trajectory-Aware Evidence Coordination for Multi-Step Visual RAG}

\author{%
Yalun Wu$^{1}$,\quad Bingzhou Wang$^{2}$,\quad Boyang Wang,\quad Peiying Wang,\quad Shaojie He, \\
\bfseries Yunhan Wang$^{3}$,\quad Shaozu Yuan,\quad Jiawei Wang$^{1}$ \\[5pt]
\normalfont\normalsize $^{1}$NExT++ Lab, National University of Singapore \\
\normalfont\normalsize $^{2}$University of Science and Technology of China \\
\normalfont\normalsize $^{3}$Beihang University
}

\maketitle
\lhead{Preprint}

\begin{abstract}

Multi-step visual retrieval-augmented generation (RAG) answers complex questions by repeatedly retrieving visual evidence, updating an intermediate state, and deciding whether to continue searching or answer. Yet retrieving relevant evidence does not ensure its effective use throughout the reasoning trajectory. As multi-step reasoning progresses, redundant sources occupy context capacity needed for missing evidence, observations tied to resolved requirements or unproductive searches linger in context, and visual sources are revisited with insufficient detail for fine-grained reading. We term this loss of usable evidence over a reasoning trajectory \textit{trajectory-level evidence utilization degradation}. To address it, we propose \textbf{Trajectory-Aware Evidence Coordination (TAEC)}, a training-free framework that coordinates evidence use around unresolved answer requirements. TAEC tracks these requirements in a shared trajectory state to guide which evidence enters the context, how accumulated memory is retained, and at what level of detail visual evidence is examined. Under a unified evaluation protocol on ViDoSeek, SlideVQA, and MMLongBench-Doc, TAEC achieves the best overall performance against leading training-free visual RAG baselines, with the highest average accuracy across multiple proprietary vision-language models.
These results demonstrate that aligning evidence with evolving reasoning
needs improves evidence use throughout multi-step visual RAG.

\end{abstract}

\section{Introduction}
\label{sec:intro}

Retrieval-augmented generation (RAG) has expanded beyond text-only knowledge sources to external multimodal memories containing both images and text \citep{murag2022}. Advances in vision-language retrieval now allow visual RAG systems to retrieve natural images or rendered document pages from external collections and answer questions using their visual and textual content \citep{colpali2025,visrag2025,m3docrag2024,visdom2025}. For complex questions, however, a single retrieval pass may not provide sufficient evidence: answering may require successive searches, query reformulation, and fine-grained visual inspection. Recent systems therefore extend the retrieve-then-read paradigm to multi-step interaction between a vision-language model and a visual collection \citep{vidorag2025,vragrl2025,visor2026,vimrag2026}.  Starting from the input question, the model retrieves and examines candidate sources, updates its reasoning state, and formulates follow-up queries when evidence is insufficient. This cycle continues until the model produces an answer or exhausts its interaction budget.

\begin{figure}[t]
\centering
\includegraphics[width=0.98\textwidth]{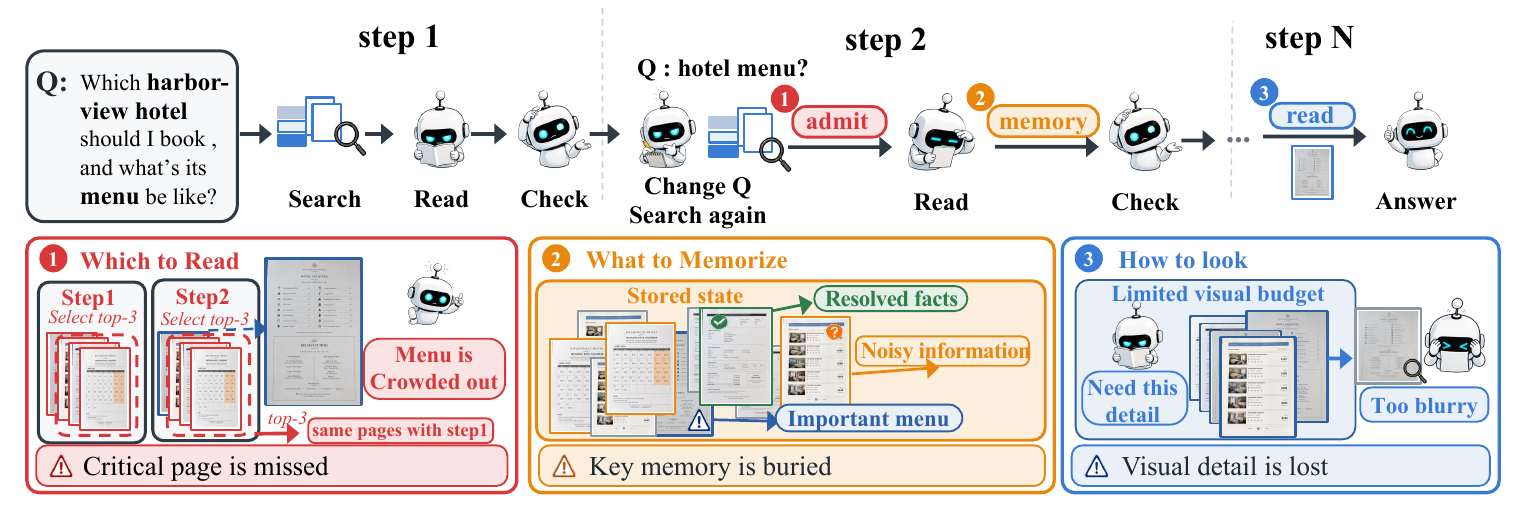}
\caption{Trajectory-level evidence utilization degradation in multi-step visual RAG.}
\label{fig:motivation}
\end{figure}

In addition to deciding what to retrieve next, multi-step visual RAG must manage both incoming evidence and accumulated observations under limited context and visual-processing capacity. Existing work supports this process through several complementary mechanisms. Methods select useful pages or fragments from retrieval results \citep{fesrag2026,megrag2026,mmagentr2_2026,adagres2025} and initiate further retrieval when the available evidence is insufficient \citep{flare2023,selfrag2024,vidorag2025,vragrl2025,fairrag2025,s2grag2026,gdprag2026}. Accumulated history is organised into summaries or memory graphs \citep{vimrag2026,magerag2026,hievirag2026}, or compressed to reduce its context footprint \citep{longllmlingua2024,memgpt2023}. Visual inspection mechanisms enable fine-grained reading through region-level zooming \citep{visualtokenscaling2025,virgo2026}, and importance-based policies determine which historical items to retain and how much visual detail to preserve \citep{visor2026}. Recent agentic systems are increasingly integrating several of these mechanisms within a single pipeline \citep{vidorag2025,vragrl2025,visor2026,vimrag2026}.

Despite these advances, existing multi-step visual RAG systems can still suffer from reduced answer accuracy over long reasoning trajectories due to ineffective use of retrieved evidence. Our analysis of ReAct \citep{react2023} on ViDoSeek shows that annotated evidence is retrieved in $74.6\%$ of trajectories involving multiple searches. As the number of searches increases, answer accuracy on these trajectories declines from $91.7\%$ for two searches to $63.8\%$ for six to eight and $52.0\%$ for nine or more. The proportion of all questions answered incorrectly despite evidence retrieval correspondingly rises from $7.9\%$ to $22.8\%$ and $28.3\%$. These results reveal a gap between retrieving relevant evidence and using it effectively over a reasoning trajectory. Accumulated observations can interfere with the use of retained evidence, while query reformulation and context updates may exclude evidence still needed to answer the original question. These processes can reduce the availability or usability of relevant evidence across reasoning steps, a phenomenon we term \emph{trajectory-level evidence utilization degradation}.

An insightful guiding principle comes from how human analysts manage evidence. They prioritise sources that address unresolved questions, distinguish established conclusions from observations that still require investigation, and revisit relevant visual material when closer inspection is needed. The common principle is to allocate limited processing resources according to what remains unresolved, rather than treating all previously relevant information as equally useful. Applying this principle to multi-step visual RAG requires coordinating three decisions. First, newly retrieved sources may be relevant to the question yet add little beyond the evidence already available, consuming context space needed for missing evidence. Second, observations tied to resolved subproblems or unproductive search branches may persist in context and compete with information needed for subsequent reasoning. Third, retained visual sources may be presented at a level of detail insufficient for the model's current information needs. These challenges concern \emph{which evidence enters the context}, \emph{how accumulated memory is retained}, and \emph{at what level of detail visual evidence is examined}. Because all three depend on the evolving reasoning state, they should be coordinated around a shared account of unresolved answer requirements.

To address this challenge, we propose \textbf{Trajectory-Aware Evidence Coordination (TAEC)}, a training-free coordination layer for multi-step visual RAG. TAEC maintains a shared trajectory state that tracks unresolved answer requirements and is updated as evidence accumulates. Three components use this state to coordinate evidence processing. \emph{Evidence Admission} selects newly retrieved evidence according to its incremental contribution to unresolved requirements. \emph{Adaptive Memory Exposure} adjusts how long accumulated memory remains exposed and in what form, based on its current role in the trajectory. \emph{Visual Detail Allocation} distributes the visual-processing budget according to the level of detail still required from each retained image. By grounding these decisions in the same trajectory state, TAEC coordinates input, memory, and visual resources around what the model still needs to answer the question. Search planning, visual interpretation, state integration, stopping decisions, and answer generation remain the responsibility of the acting model.


We evaluate TAEC against single-pass, iterative, and multi-agent
retrieval baselines on ViDoSeek, SlideVQA, and MMLongBench-Doc.
The comparisons follow a unified protocol with a common retriever,
a shared set of backbone models, the same judge, and matched
interaction budgets. TAEC achieves the best overall performance
against leading training-free visual RAG baselines, with the
highest average accuracy. Further analyses show larger gains
over ReAct on questions with longer ReAct trajectories,
highlighting the value of requirement-aware evidence coordination
in sustaining effective evidence use throughout multi-step reasoning.

The main contributions of this paper are summarized as follows:
\begin{itemize}


\item We identify and empirically validate trajectory-level evidence utilization degradation: relevant evidence can remain available without adequately supporting the model's evolving reasoning requirements.


 \item We introduce TAEC, a training-free framework that aligns evidence admission, memory exposure and visual detail allocation through a shared state of unresolved answer requirements, so that the evidence placed before the acting model keeps matching what its reasoning still needs as the trajectory grows.
 

\item We evaluate TAEC against single-pass, iterative, and
multi-agent retrieval baselines on three benchmarks under a
unified protocol. TAEC achieves the highest average accuracy
in comparisons with leading training-free visual RAG methods.
Component ablations and trajectory-level analyses examine
the contributions of the three mechanisms and their
effectiveness over extended reasoning trajectories.

\end{itemize}

\section{Related Work}

\paragraph{Visual and multi-step multimodal RAG.}
Multimodal RAG extends retrieval-augmented generation beyond text-only sources to include images
\citep{murag2022}. For visually rich documents, ColPali \citep{colpali2025} learns multi-vector
representations of rendered pages, and VisRAG \citep{visrag2025} performs retrieval and generation
directly over document images. M3DocRAG \citep{m3docrag2024} supports multi-page and multi-document
question answering, while VisDoMRAG \citep{visdom2025} combines visual and textual RAG pipelines.
Multi-step systems extend this paradigm through iterative retrieval and reasoning. ViDoRAG
\citep{vidorag2025} employs a multi-agent workflow, and VRAG-RL \citep{vragrl2025} learns retrieval
and visual-perception actions through reinforcement learning.

\paragraph{Adaptive retrieval and evidence selection.}
Adaptive RAG methods regulate when and how additional information is retrieved. Self-RAG
\citep{selfrag2024} and FLARE \citep{flare2023} use self-reflection and generation confidence,
respectively, to guide retrieval. Adaptive-RAG \citep{adaptiverag2024} selects retrieval strategies
according to question complexity, and DeepRAG \citep{deeprag2026} combines iterative query
decomposition with adaptive retrieval decisions. CRAG \citep{crag2024} evaluates retrieval quality
to guide corrective actions. Evidence sufficiency and information gaps also guide further queries
and planning in S2G-RAG \citep{s2grag2026}, FAIR-RAG \citep{fairrag2025}, and GDP-RAG
\citep{gdprag2026}. Beyond retrieval decisions, GRO-RAG \citep{grorag2026} combines relevance-
and redundancy-aware source selection with gradient-based document reranking. AdaGReS
\citep{adagres2025} balances relevance and redundancy when selecting chunks under a token budget.

\paragraph{Context and memory management.}
Effective evidence use depends on how information is retained and presented, and can be affected
by its position in long contexts \citep{liu2024lost}. LongLLMLingua \citep{longllmlingua2024}
compresses prompts, while MemGPT \citep{memgpt2023} manages information across memory tiers.
Recent agentic visual RAG systems incorporate these concerns into multi-step reasoning.
VISOR \citep{visor2026} preserves accumulated findings in a structured evidence space and
reconstructs context through a sliding window and repeated reminders of the original query.
VimRAG \citep{vimrag2026} uses semantic priority, graph dependencies, and temporal decay to
select visual memories and allocate their resolution. MAGE-RAG \citep{magerag2026} constructs
query-specific evidence subgraphs under explicit budgets.
TAEC focuses on the criteria governing evidence admission, memory exposure, and visual detail
allocation. New sources are selected according to requirement coverage and redundancy, while
historical observations are presented according to their resolution status and branch outcomes.
Visual allocation additionally considers estimated detail demand and context pressure alongside
memory relevance. These mechanisms form a training-free coordination layer
around the acting model.

\section{Trajectory-Aware Evidence Coordination}
\label{sec:method}

In multi-step retrieval, each step in a text or vision-language agent may build on a subset
of earlier observations. These dependencies can be represented as a directed acyclic graph,
whose nodes represent searches and their associated observations and whose edges indicate
dependencies on earlier findings
\citep{flare2023,selfrag2024,deeprag2026,s2grag2026,vidorag2025,vragrl2025,visor2026,vimrag2026}.
TAEC operates on this trajectory graph, denoted by $G_t$ at step $t$, with each memory item
indexed by its originating node. The observations and dependencies recorded in the graph
provide the basis for assessing requirement resolution, tracing supporting evidence,
and identifying abandoned branches.

\subsection{Unified Formulation}
\label{sec:taec_formulation}

Let $q$ denote the input question, $\mathcal{P}_t$ the visual candidates retrieved at step $t$,
$G_t$ the trajectory graph accumulated so far, and $M_t$ the cross-step multimodal memory.
The information needed to answer $q$ is represented by a compact requirement set
$\mathcal{U}=\{(u,w_u)\}$, where $u$ denotes an answer requirement and $w_u$ its importance.
Let $\bar w_{u,t}$ denote the weight requirement $u$ carries at step $t$, reflecting how far it
is already supported by the evidence in context. The resulting requirement state
$\mathcal{U}_t=\{(u,\bar w_{u,t})\}$ emphasises \emph{unresolved requirements}.
All three TAEC components use this shared state as a common basis for their decisions.

At each step, the acting model receives newly admitted sources, rendered historical observations, and retained visual memories. Admission determines which sources enter the context, while memory exposure and visual allocation affect the availability and interpretability of accumulated evidence. These decisions jointly shape the evidence presented to the model under limited processing budgets. The corresponding configuration is expressed as
\begin{equation}
\mathcal{C}_t
=
\left(
S_t,\;
\{E_{m,t}\}_{m\in M_t},\;
\{p_{i,t}\}_{i\in V_t}
\right),
\qquad
|S_t|\le K,\quad
\mathrm{ctx}(E_t)\le L_t,\quad
\sum_{i\in V_t}p_{i,t}\le B_t ,
\label{eq:taec_coordination}
\end{equation}
where $S_t\subseteq\mathcal{P}_t$ is the set of newly admitted sources and $E_{m,t}$ specifies
the exposure of historical memory $m$. The set $V_t$ contains retained historical images,
with $p_{i,t}$ denoting the visual capacity assigned to image $i$.
Here, $E_t=\{E_{m,t}\}_{m\in M_t}$, and $\mathrm{ctx}(E_t)$ denotes the context length
of the rendered memory. The budgets $K$, $L_t$, and $B_t$ constrain the number of admitted
sources, the memory context length, and the total visual capacity for retained images, respectively.

\begin{figure}[t]
\centering
\includegraphics[width=0.99\textwidth]{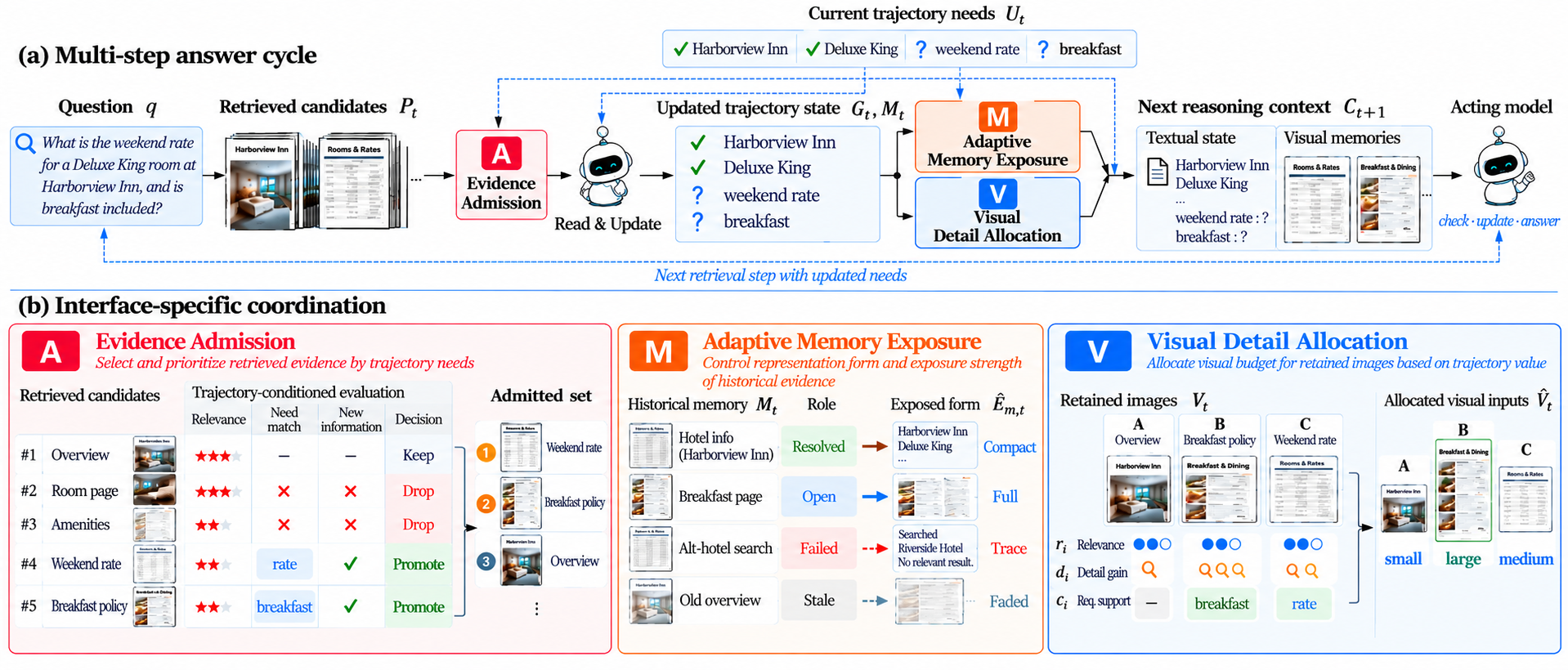}
\caption{Overview of Trajectory-Aware Evidence Coordination. At each step, the shared requirement
state guides evidence admission, memory exposure, and visual detail allocation.}
\label{fig:taec_pipeline}
\end{figure}

Figure~\ref{fig:taec_pipeline} illustrates one step of this process. The acting model remains
responsible for search planning, visual interpretation, state integration, stopping decisions,
and answer generation. Implementation details and hyperparameter settings are provided in
Appendix~\ref{app:impl}.

\subsection{Evidence Admission}
\label{sec:taec_admission}

Evidence Admission selects newly retrieved sources for inclusion in the context.
Retrieval relevance measures how closely a candidate matches the query but does not account
for information already covered by selected sources. Consequently, several highly ranked
sources may occupy input capacity with redundant information.

For candidate $i$, let $b_i$ denote its retrieval relevance and $a_{iu}$ its support for
requirement $u$. Let $C_u(S)$ denote the coverage of requirement $u$ provided by a selected
set $S$, and $\operatorname{sim}(i,j)$ the redundancy between candidates $i$ and $j$.
The admission objective is formulated as
\begin{equation}
S_t^{*}
=
\arg\max_{\substack{S\subseteq\mathcal P_t\\|S|\le K}}
\left[
\alpha_A
\sum_{(u,\bar w_{u,t})\in\mathcal U_t}
\bar w_{u,t} C_u(S)
+
\beta_A
\sum_{i\in S}b_i
-
\eta_A
\sum_{\substack{i,j\in S\\i<j}}
\operatorname{sim}(i,j)
\right].
\label{eq:admission}
\end{equation}

The three terms reward coverage of unresolved requirements and retrieval relevance while
penalising redundancy among selected sources. Their relative contributions are controlled by
$\alpha_A$, $\beta_A$, and $\eta_A$, respectively. A greedy procedure approximates the solution,
retaining the top-ranked retrieval result as an anchor to preserve the strongest retrieval signal.

\subsection{Adaptive Memory Exposure}
\label{sec:taec_memory}

Adaptive Memory Exposure determines how accumulated observations are retained and presented.
Some observations continue to support unresolved requirements, while others concern resolved
questions, duplicate existing information, or originate from unproductive searches.
TAEC therefore adjusts both memory persistence and the representation of observations
as reasoning progresses.

Existing work manages information through hierarchical organisation, graph-based evidence
selection, and energy-based memory prioritisation
\citep{memgpt2023,raptor2024,graphrag2024,mg2rag2026,magerag2026,vimrag2026}.
TAEC uses the relevance scores maintained by the memory backend and introduces an exposure
rule that combines granularity-dependent persistence with observation rendering conditioned
on the current trajectory role.
For a historical memory item $m$, let $g_m$ denote its information granularity,
$\Delta t_m$ its age, $h_m$ its stored observation, and $s_{m,t}$ its current role
in the reasoning trajectory. TAEC defines its exposure as
\begin{equation}
E_{m,t}
=
\left(
e^{-\lambda_{g_m}\Delta t_m},
\;
\mathcal{R}(h_m,s_{m,t})
\right).
\label{eq:memory_exposure}
\end{equation}

The first component modulates memory persistence through $\lambda_{g_m}$, allowing the decay rate to vary with information granularity.

The second component controls how textual observations are presented to the acting model.
The rendering function $\mathcal{R}$ preserves active or uncertain observations in full,
condenses resolved observations into concise statements of established facts, and represents
failed branches with brief terminal traces. The presentation thus adapts to the reasoning
state while the stored trajectory remains unchanged. The role $s_{m,t}$ captures
resolution status, branch outcome, recency, and repeated exposure.

\subsection{Visual Detail Allocation}
\label{sec:taec_visual}

Visual Detail Allocation determines the level of detail available when retained images are read.
Retaining a relevant image does not ensure that its content can be reliably interpreted,
particularly as visual memories compete with a growing reasoning context.
The benefit of additional visual detail also varies across images. TAEC therefore separates
image relevance from the visual capacity assigned to each retained image.

Let $V_t$ denote the set of historical images selected by the underlying memory policy,
and let $r_{i,t}$ denote the trajectory relevance of image $i$.
Let $d_{i,t}$ denote the estimated benefit of finer visual inspection and $c_{i,t}$
the image's support for requirements relevant to the current step.

TAEC allocates visual capacity according to
\begin{equation}
p_{i,t}
=
B_t
\left[
\kappa
\frac{r_{i,t}}
{\sum_{j\in V_t}r_{j,t}}
+
(1-\kappa)
\frac{
\exp\!\left(\tau_V r_{i,t}d_{i,t}c_{i,t}\right)
}{
\sum_{j\in V_t}
\exp\!\left(\tau_V r_{j,t}d_{j,t}c_{j,t}\right)
}
\right],
\qquad i\in V_t .
\label{eq:visual_allocation}
\end{equation}

The first term preserves the memory backend's allocation in proportion to historical relevance.
Its budget share is controlled by $\kappa\in[0,1]$, with $\kappa=1$ recovering this rule exactly.
The second term uses image relevance, estimated detail benefit, and requirement support
to allocate the remaining budget share, with $\tau_V$ controlling allocation concentration.
It favours images whose additional visual detail is expected to support current reasoning.
The total budget $B_t$ adapts to context pressure after memory exposure, reflecting
the context actually presented to the acting model.

\section{Experiments}
\label{sec:experiments}

We analyse evidence use across multi-step trajectories and evaluate TAEC through
baseline comparisons, component ablations, and cost measurements.

\subsection{Experiment Setting}
\label{sec:setup}

\paragraph{Protocol.}
We evaluate on three benchmarks with different evidence distributions and reasoning
requirements. \textbf{ViDoSeek} \citep{vidorag2025} covers single- and multi-hop
question answering over page images. \textbf{SlideVQA} \citep{slidevqa2023} focuses
on slide decks, with most questions answerable from a single slide.
\textbf{MMLongBench-Doc} \citep{mmlongbench2024} requires locating relevant evidence
within long documents; we evaluate on its 847 questions annotated as answerable.
We use four proprietary vision-language models as acting models:
Gemini-3.5-Flash, Kimi-K3, GPT-5.6-Sol and GPT-4o-mini, spanning different deployment
tiers. All models remain frozen, and no TAEC component is trained.
For comparisons within each table, all systems use the same read-only retrieval
index, the same limit on pages admitted per search, and the same interaction and
visual budgets. Outputs are evaluated together in a single batch by GPT-4.1 using
a fixed judging prompt. Appendix~\ref{app:expdetails} provides details of the index,
budgets, judge, acting-model selection and inference settings, and the source of
each reported result.

\paragraph{Systems.}
We compare TAEC with representative retrieval baselines and leading training-free
agentic RAG systems. \textbf{Vanilla} performs a single retrieval followed by
answer generation. \textbf{ReAct} \citep{react2023} follows a think--search--observe loop and retains
all retrieved images in the context. \textbf{ViDoRAG} \citep{vidorag2025} is a
leading training-free multi-agent framework with publicly available code.
\textbf{DAG agent} uses the underlying trajectory graph without the three TAEC
components. \textbf{TAEC} augments this framework with evidence admission,
adaptive memory exposure and visual detail allocation. We additionally compare
against \textbf{M3RAG} \citep{m3rag2026} using our reimplementation, as its code
is not publicly available; this comparison is reported in
Appendix~\ref{app:m3rag}.

\paragraph{Metrics.}
We report answer accuracy throughout. To analyse evidence availability and use,
we additionally report \emph{coverage}, the proportion of questions for which
at least one annotated evidence page (a \emph{gold page}) enters the context at
any point during the trajectory. We define \emph{use} as answer accuracy on this
subset. Appendix~\ref{app:gaindecomp} details the decomposition used for comparisons
across systems.

\subsection{Evidence Utilisation over Long Trajectories}
\label{sec:diagnosis}


We analyse ReAct trajectories on ViDoSeek with Gemini-3.5-Flash, where all
retrieved images remain in the context as reasoning proceeds.

\needspace{3.4in}
\begin{wrapfigure}{r}{0.47\textwidth}
\centering
\includegraphics[width=0.45\textwidth]{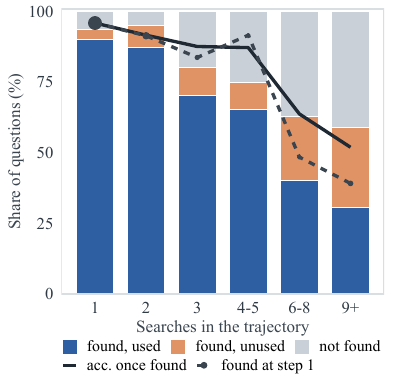}
\caption{ReAct on ViDoSeek with Gemini-3.5-Flash, grouped by the number of searches.
Marker area is proportional to the number of questions in each group.}
\label{fig:evidence_loss}
\end{wrapfigure}
Among trajectories that retrieve a gold page, answer accuracy decreases as the
number of searches increases (solid line in Figure~\ref{fig:evidence_loss}).
The orange bar segments show the proportion of all questions in each group
for which a gold page is retrieved but the final answer is incorrect.
Table~\ref{tab:depth_loss} reports the complete breakdown.

To examine this pattern when evidence is available from the first step, we
restrict the analysis to questions whose gold page appears among the first
five retrieved pages (dashed line). Accuracy is $96.0\%$ for trajectories
ending after one search and $39.2\%$ for those involving nine or more searches.

A controlled intervention on the DAG agent further examines the effect of
context growth while holding the questions and retrieved pages fixed.
Adding $32$K neutral text tokens reduces accuracy by $5.2$ percentage points,
with the gold pages retained and their visual presentation unchanged
(Appendix~\ref{app:interference}). Together, these findings indicate that
evidence availability alone does not ensure effective use, motivating
trajectory-level evidence coordination.

\subsection{Main Results}
\label{sec:main}

Table~\ref{tab:main} compares methods using the same retriever, interaction
budget, and judge on identical question sets. Among the methods in this table,
TAEC achieves the highest accuracy in all twelve settings. With
Gemini-3.5-Flash, it outperforms ViDoRAG \citep{vidorag2025}, a leading
training-free multi-agent framework with publicly available code, by $7.8$,
$11.1$ and $18.6$ percentage points on ViDoSeek, SlideVQA and MMLongBench-Doc,
respectively. Its corresponding gains over our adapted implementation of M3RAG
\citep{m3rag2026} are $5.0$, $2.1$ and $2.5$ percentage points. Across all four
acting models, TAEC leads in ten of the twelve comparisons, with both
exceptions on MMLongBench-Doc (Appendix~\ref{app:m3rag}).

\begin{table}[!ht]
\centering
\caption{Accuracy (\%).
\textbf{Bold} and \underline{underlining} mark the best and second-best
results per column; \textbf{Avg.} is the mean across twelve settings.
MMLongBench-Doc uses 847 answerable questions;
full-set results appear in Appendix~\ref{app:answerable}. Supplementary results for M3RAG,
RL-trained agents, and open-weight backbones appear in Appendices~\ref{app:m3rag},
\ref{app:rl}, and~\ref{app:open}, respectively.}
\label{tab:main}
\vspace{3pt}
\small
\setlength{\tabcolsep}{2.6pt}
\renewcommand{\arraystretch}{1.12}
\newcommand{\hd}[1]{{\scriptsize #1}}
\begin{tabular}{l *{4}{c} @{\hskip 7pt} *{4}{c} @{\hskip 7pt} *{4}{c} @{\hskip 7pt} c}
\toprule
& \multicolumn{4}{c}{\textbf{ViDoSeek}} & \multicolumn{4}{c}{\textbf{SlideVQA}}
& \multicolumn{4}{c}{\textbf{MMLongBench-Doc}} & \\
\cmidrule(lr){2-5}\cmidrule(lr){6-9}\cmidrule(lr){10-13}
\textbf{System} & \hd{Gemini} & \hd{Kimi} & \hd{GPT-5.6} & \hd{4o-mini}
& \hd{Gemini} & \hd{Kimi} & \hd{GPT-5.6} & \hd{4o-mini}
& \hd{Gemini} & \hd{Kimi} & \hd{GPT-5.6} & \hd{4o-mini} & \textbf{Avg.} \\
\midrule
Vanilla   & 76.7 & 81.8 & 82.0 & 59.2 & \underline{79.6} & 78.0 & 79.3 & \underline{61.9} & 40.1 & 31.8 & 38.1 & \underline{15.6} & 60.3 \\
ReAct     & 79.5 & 82.1 & 82.1 & 56.1 & 77.2 & \underline{82.0} & \underline{81.0} & 59.3 & 42.4 & 36.5 & \underline{39.1} & 15.2 & 61.0 \\
ViDoRAG   & 79.7 & 80.8 & 81.3 & \underline{65.8} & 72.5 & 74.5 & 78.3 & 58.7 & 31.5 & 37.2 & 37.8 & 14.1 & 59.4 \\
DAG agent & \underline{80.1} & \underline{84.8} & \underline{84.1} & 65.7 & 76.7 & 80.9 & 77.7 & 56.6 & \underline{44.5} & \underline{43.9} & 38.3 & 14.3 & \underline{62.3} \\
\midrule
\rowcolor{tabhl}
\textbf{TAEC} & \textbf{87.5} & \textbf{88.1} & \textbf{87.0} & \textbf{71.0} & \textbf{83.6} & \textbf{83.8} & \textbf{82.2} & \textbf{62.0} & \textbf{50.1} & \textbf{45.8} & \textbf{42.6} & \textbf{17.7} & \textbf{66.8} \\
\bottomrule
\end{tabular}
\end{table}

Three further observations emerge. With Gemini-3.5-Flash, TAEC's gains over
Vanilla vary across benchmarks: $4.0$ percentage points on SlideVQA, compared
with $10.0$ on MMLongBench-Doc and $10.8$ on ViDoSeek. ReAct underperforms
Vanilla in four of the twelve settings, showing that iterative retrieval alone
does not consistently improve accuracy. Gains from coordination also vary
across acting models. Gemini-3.5-Flash benefits more than Kimi-K3 and
GPT-5.6-Sol despite their similar accuracy on single-search questions with a
gold page available (Appendix~\ref{app:frontier}). For Gemini-3.5-Flash, adding
the three components to the DAG agent improves accuracy by $7.4$, $6.9$ and
$5.6$ percentage points on ViDoSeek, SlideVQA and MMLongBench-Doc,
respectively.

\subsection{Component Ablation}
\label{sec:components}

\begin{wrapfigure}{r}{0.47\textwidth}
\centering
\includegraphics[width=0.45\textwidth]{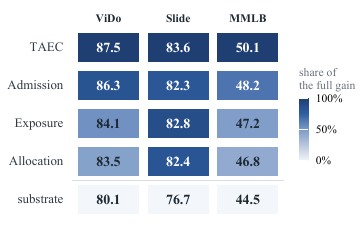}
\caption{Component ablation with Gemini-3.5-Flash.
Values show accuracy (\%); colour indicates gains over the DAG agent,
normalised by TAEC's gain on each benchmark.}
\label{fig:ablation}
\end{wrapfigure}

We add each component individually and all three jointly to the DAG agent,
evaluating their contributions on all three benchmarks with Gemini-3.5-Flash.
Figure~\ref{fig:ablation} summarises the results, with detailed values in
Table~\ref{tab:components}.

\Needspace{18\baselineskip}

Each component improves accuracy over the DAG agent on all three benchmarks,
achieving approximately $41\%$--$88\%$ of TAEC's gain. The full system achieves
the highest accuracy throughout.
Admission yields the largest individual gains on ViDoSeek and MMLongBench-Doc,
achieving $6.2$ of the full $7.4$ percentage-point gain and $3.7$ of $5.6$,
respectively. This pattern is consistent with the importance of page selection:
most ViDoSeek questions are answered after one search, while MMLongBench-Doc
requires locating evidence within long documents. Memory exposure and visual
allocation mainly act on subsequent reasoning steps.
On SlideVQA, the three individual configurations differ by at most $0.5$
percentage points, and each achieves more than $80\%$ of the full gain.

\par\WFclear

\subsection{Analysis}
\label{sec:supplementary}

\begin{figure}[t]
\centering
\includegraphics[width=\textwidth]{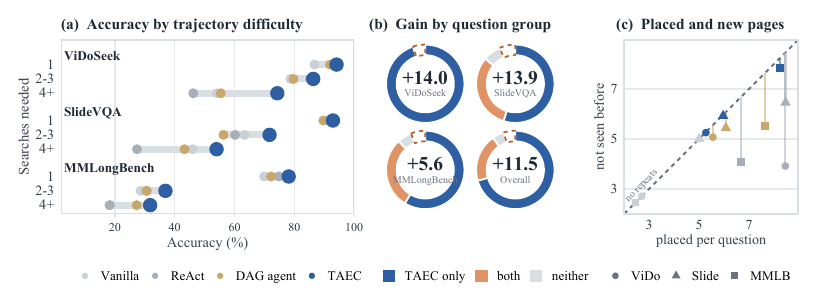}
\caption{Analysis with Gemini-3.5-Flash (Table~\ref{tab:main}).
(a) Accuracy by ReAct search count (Table~\ref{tab:multistep_main}).
(b) TAEC's accuracy gain over the DAG agent, decomposed by coverage group
on questions with multiple ReAct searches. Centres show net gains
(percentage points); hollow sectors indicate negative contributions.
(c) Mean admitted and distinct pages per question.
MMLongBench-Doc coverage uses answerable questions with annotated evidence pages.}
\label{fig:results}
\end{figure}

\paragraph{Trajectory depth.}
Figure~\ref{fig:results}(a) groups questions by the number of searches performed
by ReAct, using identical question subsets for all systems
(Table~\ref{tab:multistep_main}, Appendix~\ref{app:tables}).
TAEC's advantage over ReAct is larger in groups with more searches.
In the group with at least four searches, ReAct achieves $46.2\%$, $27.4\%$
and $18.2\%$ accuracy on ViDoSeek, SlideVQA and MMLongBench-Doc, respectively,
compared with $74.3\%$, $54.0\%$ and $31.8\%$ for TAEC.
ReAct ranks last among the four systems in each of these groups.

\paragraph{Coverage and use.}
Figure~\ref{fig:results}(b) decomposes TAEC's accuracy gains over the DAG agent
on the fixed subsets where ReAct performs multiple searches.
The gains are $14.0$, $13.9$ and $5.6$ percentage points on ViDoSeek, SlideVQA
and MMLongBench-Doc, respectively. The largest positive contributions come
from questions for which only TAEC retrieves a gold page: $14.8$, $8.8$ and
$3.8$ percentage points. On the 115 such questions in ViDoSeek, accuracy is
$22\%$ for the DAG agent and $86\%$ for TAEC.
Questions for which both systems retrieve a gold page contribute $-0.6$,
$5.0$ and $1.9$ percentage points, respectively. Within these shared subsets,
TAEC is $8.8$ and $5.0$ percentage points higher on SlideVQA and MMLongBench-Doc,
and level on ViDoSeek, where the two differ by three of $339$ questions. These groups are defined by ReAct's search count; TAEC
itself answers $93.5\%$ of ViDoSeek questions after one search.
Appendix~\ref{app:gaindecomp} provides the full four-group decomposition and
explains why post-hit accuracy cannot be compared directly across systems.

\paragraph{Context composition.}
All systems use the same limit on pages admitted per search.
Figure~\ref{fig:results}(c) compares the mean numbers of admitted and distinct
pages per question. On ViDoSeek, only $46.3\%$ of ReAct's admitted pages are
distinct, so more than half of its page inputs are repeats.
The DAG agent reduces repetition by tracking previously retrieved pages.
With TAEC, distinct pages account for $99.6\%$, $99.3\%$ and $95.2\%$ of
page admissions on ViDoSeek, SlideVQA and MMLongBench-Doc, respectively.

\paragraph{Retrieval effort.}
On the fixed subsets where ReAct performs multiple searches, TAEC achieves
higher coverage with fewer searches. Across the three benchmarks, ReAct
averages $3.00$--$6.00$ searches per question, compared with $1.28$--$2.58$
for TAEC. On ViDoSeek, coverage increases from $74.6\%$ with ReAct to
$90.8\%$ with TAEC, with the same ordering on the other two benchmarks.
The DAG agent achieves coverage of $68.4\%$, $59.3\%$ and $40.2\%$ on the
corresponding subsets (Table~\ref{tab:matched}, Appendix~\ref{app:tables}).

\subsection{Efficiency}
\label{sec:cost}

By reducing redundant inputs and repeated evidence transmission, TAEC can lower
interaction overhead. On ViDoSeek with Gemini-3.5-Flash, it transmits an
average of $0.61$~MB of context per question, compared with $0.87$~MB for the
DAG agent and $3.73$~MB for ReAct. It also requires fewer model calls than both
and reduces cumulative image transmissions from $92.8$ per question for ReAct
to $14.7$ (Table~6, Appendix~C). These savings accompany higher answer
accuracy.

Additional analyses examine alternative retrievers (Appendix~K), search
depth (Appendix~L), and performance across backbones, including open-weight
and RL-trained models (Appendices~I and~J).
Further baseline comparisons and an audit of the evaluation protocol are
reported in Appendices~D and~N, respectively.

\subsection{Limitations}
\label{sec:limitations}

TAEC coordinates the evidence presented to the acting model, which remains
responsible for assessing evidence sufficiency and controlling the reasoning
trajectory. Its effectiveness therefore depends on the model's existing
agentic capabilities (Appendix~\ref{app:frontier}). As a training-free evidence coordination
layer, TAEC cannot fully compensate for deficiencies in these capabilities;
policy training remains a complementary direction for improving the acting
model. With Qwen2.5-VL-7B, single-pass retrieval outperforms the evaluated
multi-step configurations on SlideVQA (Appendix~\ref{app:open}). With the
smaller Qwen3-VL-4B, TAEC outperforms single-pass retrieval, indicating that
this limitation is not universal among small or open-weight models.
The main comparisons are limited to a single retriever family, with
alternative retrievers examined in Appendix~\ref{app:retrievers}. All three benchmarks
consist of English document collections.

\section{Conclusion}

We identify \emph{trajectory-level evidence utilization degradation} in multi-step visual RAG and propose TAEC, a training-free framework that tracks unresolved answer requirements to coordinate evidence admission, memory exposure, and visual detail allocation. On ViDoSeek, SlideVQA, and MMLongBench-Doc, TAEC achieves the highest average accuracy against leading training-free baselines across multiple proprietary vision-language models. The gains are more pronounced on longer trajectories, supporting requirement-aware evidence coordination as an effective approach to sustaining evidence use in multi-step visual RAG.


\bibliographystyle{iclr2027_conference}
\bibliography{references}

\appendix

\section{Implementation Details}
\label{app:impl}

This section documents the substrate and the three components as they run in the evaluated code, and
maps the panels of Figure~\ref{fig:taec_pipeline} onto that code. Simplifications relative to the formulation in
Section~\ref{sec:method} are noted where they occur. Every coefficient, decay
rate and threshold below is the default of the released implementation, unchanged across all
benchmarks, backbones and retrievers; the exact values are in the released code rather than restated
here.

\subsection{Substrate: the DAG agent}
\label{app:impl_substrate}

\paragraph{The trajectory graph.}
The trajectory is built by the acting model itself, one node per action. A \emph{search node} records
the query the model issued, the identifiers of the earlier nodes it declared this search to follow
from, and, once the model has read the returned pages, the summary it wrote about them and the pages it
marked for retention. An \emph{answer node} terminates a path and records the final answer and the
nodes it rests on. Edges are therefore claims of dependence made by the model at the moment it acts:
a search issued to refine an earlier finding names that finding as its parent, while a search that
opens an independent line names the root, so sibling subtrees correspond to independent lines of
inquiry and a chain corresponds to one line being pursued. Every visual memory is stored under the
node that admitted it, so the graph indexes both what was concluded and where the evidence for it came
from, and a node's position in the graph, its depth, its number of children and whether its subtree
ever reached an answer, is read by later steps when they assess how much a memory still contributes.

\paragraph{Tool interface and loop.}
The acting model sees three tools. \texttt{add\_search\_node(id, parent\_ids, query)} issues one
retrieval call and returns the admitted pages as images, each preceded by a \texttt{Picture $k$}
label, together with a fixed instruction to summarise. \texttt{summarize\_and\_memorize(summarize,
memorize)} returns a one-to-three-sentence summary and an optional list of useful pictures, each with
a self-reported \texttt{priority\_score} in $\{1,\dots,5\}$. \texttt{add\_answer\_node(parent\_ids,
answer)} ends the trajectory. Each search must be followed by a summarise call before the next search
is accepted; after every summarise the prompt is rebuilt from scratch. The prompt contains the system
instruction, the question with any runtime hints, the full action graph serialised as JSON (node ids,
parents, queries, summaries), and a \emph{multimodal memory} block of re-presented images. Retrieval
returns five candidates per query; pages already shown earlier in the trajectory are removed from new
results. The loop runs for at most 20 model calls;
if the model has not called the answer tool, a final call with a forced-answer instruction is issued,
and if that still yields no tool call, a plain-text question is asked over the same context.

\paragraph{Energy-based memory.}
Every picture the model marked useful is stored with its node id, granularity (page or region) and
priority. Before each call, the memory is scored with a graph energy of the form used by graph-structured
agent memories \citep{memgpt2023,raptor2024,graphrag2024,mg2rag2026,magerag2026,vimrag2026},
\begin{equation}
\Omega_i=\left(\tfrac{p_i}{5}\right)\,(1+\mathrm{outdeg}(m_i))\,e^{-\lambda_{g_i}\Delta t_i}
+\gamma\sum_{c\in\mathrm{children}(m_i)}\bar\Omega_c ,
\label{eq:impl_energy}
\end{equation}
where $m_i$ is the node that admitted image $i$, $\Delta t_i$ its age in nodes, $\gamma$ the
child-feedback weight, and $\lambda_{g}$ a granularity-dependent decay rate that is zero for
entity-level text and largest for whole pages. The highest-energy images are re-presented under the
memory pixel budget of Appendix~\ref{app:expdetails}, and that budget is scaled up as the textual
prompt grows. Three properties of this instantiation belong to the components of
Section~\ref{sec:method} rather than to the backend: the decay rate $\lambda_g$ is granularity-dependent
rather than global, the priority $p_i$ is the measured information gain of an image rather than a
self-reported score, and the total budget follows context pressure instead of being fixed. Two runtime hints are derived from summaries by keyword cues:
a summary with a failure cue (``not found'', ``unclear'', and their Chinese equivalents) marks its
query as a dead end, and the next prompt carries a typed hint (no result / partial / unclear) naming
that query; a summary with a high-confidence cue and a number sets an early-stop flag that asks the
model to answer next. These cues are also what adaptive memory exposure reads (below). Images are JPEG re-encoded at
most $300$K pixels and at most 38~KB each. \emph{DAG agent} is the same agent reduced to this structure. It reasons over the same trajectory
graph and writes the same summaries, and it re-presents its memories with uniform relevance, at a
visual budget that does not move with context pressure, over candidate pages ranked by the retriever
alone; the trajectory's branch outcomes stay in the graph and are not turned into signals that shape
later prompts.

\subsection{Requirement state $\mathcal{U}_t$}
\label{app:impl_state}

In the evaluated configuration the requirement set is extracted once per question by a single call
to GPT-4.1-mini at temperature 0 (``extract 2--4 information slot keywords needed to answer the
question''), cached, and truncated to four slots with uniform weight. Coverage $C_u(S)$ is updated
inside the admission greedy as slots become covered by admitted pages. In the runs reported here the slot
weights stay uniform; the variant in which the acting model reports resolution status at every
summarise call is implemented behind \texttt{VRAG\_AGENT\_SEMANTIC\_STATE} and is not enabled.
Memory exposure takes its resolution signal from the summary cues of
Appendix~\ref{app:impl_substrate}, and admission propagates its slot-coverage score into the
detail term of visual allocation. This is one call per question in addition to the acting model's own calls.

\subsection{Evidence admission}
\label{app:impl_admission}

Admission replaces the retriever's top-5 with a greedy selection over an over-sampled pool of
$5\times5=25$ candidates (capped at 50). For candidate $i$ with retriever score $b_i$ (min--max
normalised within the pool) and page caption $\mathrm{cap}_i$:
\begin{itemize}
\item slot support $a_i=\frac{1}{|\mathcal U|}\sum_{u\in\mathcal U}\mathbb 1[u\subseteq\mathrm{cap}_i]$,
the fraction of slot keywords that occur as substrings in the caption;
\item redundancy $\operatorname{sim}(i,S)=\max_{j\in S}\mathrm{Jaccard}(\mathrm{tok}_i,\mathrm{tok}_j)$
over caption tokens (CJK characters plus Latin words and numbers).
\end{itemize}
The retriever's top-1 is always kept as an anchor. The remaining slots are filled greedily by the
marginal gain of \eqref{eq:admission}, $a_i$ standing in for the coverage term and $b_i$ for the
relevance term. The admitted set is then passed through the substrate's seen-page filter and rendered
at the standard per-image resolution. Slot
scores are recorded as metadata on each admitted image and are propagated to visual detail
allocation (\texttt{VRAG\_UEP\_PROPAGATE\_SLOT\_SCORE}) in the runs reported here. Captions are the
VLM-generated page captions described in Appendix~\ref{app:expdetails}. No image is cropped in
the main runs.

\subsection{Adaptive memory exposure}
\label{app:impl_memory}

The persistence term of \eqref{eq:memory_exposure} is the granularity-specific decay
$e^{-\lambda_{g}\Delta t}$ of \eqref{eq:impl_energy}. The representation term $\mathcal R$ is
applied to the action-graph JSON at prompt-build time and never modifies the stored graph. The root node and the most recent nodes are always rendered in full. Every older node is classified
from its summary by the confidence cues: \emph{resolved} if a confidence cue is present and no failure
cue is; \emph{failed} if a failure cue is present or the node's query is in the dead-end list;
\emph{active} otherwise. Resolved nodes are rendered as one canonical sentence, the first sentence of
the summary carrying a confidence cue; failed nodes are rendered as a short terminal trace naming the
abandoned query; active nodes are rendered in full.

\subsection{Visual detail allocation}
\label{app:impl_visual}

Allocation runs only where historical images are re-presented; images retrieved at the current step
are read immediately after admission and are not part of $V_t$.

Allocation acts only when historical images are re-presented, which in the main runs means
trajectories with at least two searches (three graph nodes); shorter trajectories fall through to the
substrate's allocation. Its inputs are the memory energies $r_i=\Omega_i$ from
\eqref{eq:impl_energy}, a detail proxy $d_i$, and requirement support $c_i$. In the evaluated runs $d_i$ is maximal for region crops and, for pages, the caption length normalised
across the corpus, with the admission slot score folded into it; $c_i$ is $1$ throughout, so the
requirement signal reaches this component through $d_i$ rather than through $c_i$. The budget follows the same pressure-scaled form as the substrate's, with an
earlier onset. Which images are shown is decided by $r_i$ alone (top-5); only their
pixel shares change. Within each granularity group, image $i$ receives
$p_i=B_t^{g}\bigl[\kappa\,r_i/\!\sum_j r_j+(1-\kappa)\,\mathrm{softmax}_j(\tau\,r_jd_jc_j)_i\bigr]$,
\eqref{eq:visual_allocation} restricted to the group and clipped to the per-image resolution
limits; when region crops are present they receive the larger share of $B_t$. The \emph{no pressure} ablation fixes
$B_t=B_0$; the \emph{no detail} ablation sets $d_i=1$.

\subsection{Composition and cost}
\label{app:impl_composition}

The three components override three disjoint methods of the substrate class (retrieval handling,
prompt construction, memory rendering) and are composed by multiple inheritance in the order
A~$\rightarrow$~M~$\rightarrow$~V; the single-component variants of Table~\ref{tab:components}
enable one of them at a time. TAEC adds exactly one model call per question (slot extraction) and no calls
per step; all other computation is string matching and arithmetic on the stored graph.

\subsection{Reading the overview figure}
Panel (a) of Figure~\ref{fig:taec_pipeline} is one iteration of the loop above: retrieval returns
$\mathcal P_t$ (25 over-sampled candidates), admission reduces them to $S_t$ (five pages), the acting
model reads them and calls summarise, producing the node that updates $G_t$ and $M_t$; before the next
call, memory exposure renders $G_t$ and visual allocation budgets the re-presented images of $M_t$.
Panel (b), left, shows the admission objective: the greedy keeps the top retrieval result, then prefers
pages whose captions cover uncovered slots and penalises caption overlap with already-admitted pages.
Panel (b), middle, shows the four rendering actions (full, compact, trace, faded): \emph{full} and
\emph{compact} are the active and resolved rules, \emph{trace} is the retired rule, and \emph{faded}
is the page-level decay acting on the image side. Panel (b), right, shows the budget
split among retained images; in the evaluated runs the ``supports current needs'' row is constant
($c_i=1$) and the ``detail gain'' row is the caption-length proxy combined with the slot-coverage
score admission passes on.

\subsection{Baselines}
\textbf{Vanilla} retrieves once with the question, shows the five pages at $300$K pixels each and
asks for the answer in one call. \textbf{ReAct} exposes \texttt{search(query)} and
\texttt{answer(response)}; every observation (the five retrieved pages and their captions) stays in
the message history; the $t$-th call therefore carries all $5t$ images. Under a budget $B$ the pixels of
all images in context are split uniformly, $\min(300\mathrm K, B/n)$ each, re-encoded at every call;
$B{=}0$ means no cap. A repeated query returns a notice instead of results. The same 20-call ceiling,
forced-answer call and plain-text fallback apply. \textbf{ViDoRAG} is run from its public
implementation against the shared index with the same backbone and judge.

\section{Experimental Details}
\label{app:expdetails}

This section records the full protocol behind Section~\ref{sec:experiments}; the implementation of
the systems themselves is in Appendix~\ref{app:impl}.

\paragraph{Datasets.}
\textbf{ViDoSeek} \citep{vidorag2025} contains 1,142 questions over 290 PDF documents rendered as page
images, each question annotated with exactly one evidence page, which we call its \emph{gold page}; 645 questions are single-hop and 497
multi-hop, and by source type 730 target two-dimensional layouts, 175 tables, 157 charts and 80 running
text. \textbf{SlideVQA} \citep{slidevqa2023} is the 2,215-question test split over slide decks.
\textbf{MMLongBench-Doc} is 1,091 questions over long documents, the longest of the three and the one
that most often needs evidence from more than one page. We use the released questions and reference
answers of all three unchanged. On MMLongBench-Doc every table of the main text scores the 847 questions
whose reference answer is not \emph{Not answerable}; Appendix~\ref{app:answerable} gives the full set.

\paragraph{Retrieval indices.}
All systems in a table share one retriever behind an HTTP contract that returns page identifiers and
scores only, so that the agent code is identical across retrievers. The main index is a dense
Qwen3-VL-Embedding-2B FAISS index over a pooled corpus of 36,233 pages drawn from several document
benchmarks; pooling makes retrieval harder than the per-benchmark pools used in prior work, and it is
the same pool for every system and every backbone. SlideVQA is served from its own index of the same
type. Two further retrievers appear only in the control of Appendix~\ref{app:retrievers}: a
ColQwen2.5-v0.1 \citep{colpali2025} late-interaction index over the ViDoSeek pages, and a BM25Okapi
index over one text string per page. Every system places at most five pages in the context per
search; evidence admission selects those five from an over-sampled pool of the same index
(Appendix~\ref{app:impl_admission}). Page captions,
used by evidence admission for slot matching and by visual detail allocation as a detail proxy, are VLM-generated captions produced
by us and held fixed across all runs; their file hash is recorded in every run manifest.

\paragraph{Retriever fingerprint.}
Because a search URL does not identify a retriever, every run computes a behavioural fingerprint at
start-up: eight fixed probe queries (\emph{annual revenue table}; \emph{organizational chart};
\emph{experimental results comparison}; \emph{project timeline schedule}; a Chinese safety-procedure
phrase; \emph{2024}; \emph{conclusion and future work}; \emph{figure caption}) are sent with
$k{=}5$, the returned page file names are concatenated in rank order and hashed to a 16-hex-digit
string. The search services run read-only, expose a version endpoint with the index identity and page
count, and a run refuses to start if its expected fingerprint does not match. Tables in the paper
never mix rows with different fingerprints.

\paragraph{Choice of acting models.}
The four proprietary backbones of Table~\ref{tab:main} are one model from each of the deployment
tiers at which a multi-step visual RAG system is run. Gemini-3.5-Flash is a fast, low-cost tier model
of the kind such systems are usually deployed on. Kimi-K3 is a large model with an extended reasoning
budget. GPT-5.6-Sol is its vendor's current general-purpose model over the period of these
experiments. GPT-4o-mini is a small proprietary model, and stands for the
lower end of the same market. Appendix~\ref{app:open} extends the reading to open weights served
locally. The tiers fix the selection; what the four turn out to span once they run on one substrate
is the subject of Appendix~\ref{app:frontier}.

\paragraph{Backbones and calls.}
Every proprietary backbone is called through an OpenAI-compatible chat endpoint with native tool
calling, temperature 0.3, extended thinking disabled, and no system-side caching; the acting model is
the only thing that changes between the columns of a table. The model identifier used for each
run is recorded in the run manifest released with the code.
Qwen2.5-VL-7B/3B are served locally
with vLLM under the same tool interface. Each API key runs at most three concurrent questions behind
a token bucket; a rate-limit response backs off and retries the question, and any other failure is
recorded in the result row rather than retried silently.

\paragraph{Budgets.}
At most 20 model calls per question; a question not answered within the budget
receives a forced-answer call and, failing that, a plain-text fallback, and is scored on whatever it
returns. Freshly retrieved images are sent at most 300K pixels, re-presented memory images share a
600K-pixel budget, and every image is JPEG-encoded at quality 75 and capped at 38~KB. For ReAct under
a visual budget $B$, all images in the message history are re-encoded at every call at
$\min(300\mathrm{K}, B/n)$ pixels each, where $n$ is the number of images in context.

\paragraph{Judging.}
Every reported accuracy is produced by GPT-4.1 at temperature 0 acting as a binary judge over the
question, the reference answer and the generated answer, with one fixed prompt for all systems,
backbones and benchmarks, and all numbers in the paper come from a single judging batch. The prompt
presents the three fields, asks whether the generated answer is correct, notes that it may carry
information beyond the reference, and requires the verdict as \texttt{True} or \texttt{False} inside
a tag. It states no rule beyond that, so nothing in it can favour one system's answer style, and the
judge is not tuned per system; every row in this paper is comparable to every other, including
across backbones and benchmarks.

These benchmarks were not released with a metric of this kind, and the reason recent work has moved
away from theirs is worth recording. SlideVQA \citep{slidevqa2023} is scored by exact match and
F1, and MMLongBench-Doc \citep{mmlongbench2024} by a rule-based calculator over short answers that
GPT-4o extracts from the response, matching exactly or by ANLS. Neither survives free-form agent
output, which carries the answer inside a sentence. Agentic visual RAG work therefore scores the
answer with a model: ViDoRAG \citep{vidorag2025} grades against the reference on a five-point scale
with GPT-4o and counts four or above as correct, while VRAG-RL \citep{vragrl2025} and VISOR
\citep{visor2026} take a binary verdict and report its mean as accuracy, the latter with
Qwen-max-latest. We follow that binary form and differ in the judge model. VISOR re-scores with a
second judge and reports that the substitution moves overall accuracy by less than $0.3$ points, so
the choice of judge model is not where these comparisons are settled.
One consequence should be stated plainly: our MMLongBench-Doc numbers are not that benchmark's
official generalized accuracy, and are comparable within this paper rather than against values
reported under the rule-based protocol.
The prompt is given verbatim below; the three braced fields are the only substitutions.

\begin{promptbox}[title={Judge prompt}]
\footnotesize
\begin{verbatim}
You are an expert evaluation system for a question answering
chatbot.

You are given the following information:
- the query
- a generated answer
- a reference answer

Your task is to evaluate the correctness of the generated answer.

## Query
{query}

## Reference Answer
{reference_answer}

## Generated Answer
{generated_answer}

Your response should be formatted as following:
<judge>True or False</judge>

If the generated answer is correct, please set "judge" to True.
Otherwise, please set "judge" to False.

Please note that the generated answer may contain additional
information beyond the reference answer.
\end{verbatim}
\end{promptbox}

\paragraph{Metrics.}
\emph{Accuracy} is the mean judge verdict over all questions of a benchmark, so every system is scored
on the same denominator. A question on which a system returns no answer, because its pipeline fails,
times out or ends without a response, stays in that denominator and counts as incorrect. ViDoRAG's
released evaluation instead drops such questions. We keep them, because a denominator that excludes a
system's own failures differs from system to system and rewards a pipeline for failing on the
questions it would have answered wrongly, the same selection effect that makes post-hit accuracy
incomparable across systems (Appendix~\ref{app:gaindecomp}). \emph{Trajectory hit rate} is the
fraction of questions whose gold page enters the context at any point in the trajectory: the union,
over the searches of a trajectory, of the pages the system placed before the acting model after its
own selection and deduplication. \emph{Post-hit accuracy} is accuracy on the
hit subset. \emph{Searches} counts completed retrieval calls. \emph{Context cost} sums, over all model
calls of a question including the forced-answer call, the base64 image bytes actually transmitted;
\emph{images} and \emph{model calls} are counted the same way. Wall-clock time is per question,
including retrieval and judge-free.

\paragraph{Analyses.}
Difficulty buckets in Table~\ref{tab:multistep_main} are defined by the number of searches issued by
unconstrained ReAct on the same question, so all systems face identical question sets per bucket.
The matched subsets of Table~\ref{tab:matched} are defined by the unmanaged agent alone, by whether
it stopped after one search or kept searching, so every system in a subset answers the same
questions. Their four outcome shares are taken over the questions that carry a reference page, since
a question without one cannot be scored for whether its gold page arrived; on MMLongBench-Doc that
is 532 of the 541 questions in the subset.
Figure~\ref{fig:gain_decomp}(a) uses a second, joint definition, given there.
The search-depth table (Appendix~\ref{app:searchcap}) is computed from the recorded trajectories: a
question counts at depth $k$ when it was answered using at most $k$ searches and answered correctly;
no runs are repeated.

\paragraph{Provenance.}
Every result row stores the code fingerprint, the retriever fingerprint, the complete environment,
the query string sent at every search, and the context statistics above; every batch writes a manifest
with the judge model, the index identity and page count, and the caption file hash. API runs were
executed on a workstation; the 7B/3B models and the ColQwen2.5 service ran on a single-GPU server.

\section{Full Tables and Figures for the Main Text}
\label{app:tables}

\begin{figure}[!ht]
\centering
\includegraphics[width=\textwidth]{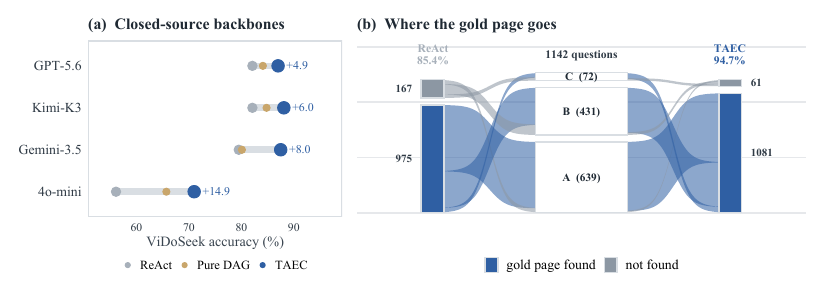}
\caption{Two appendix studies whose numbers are in the tables named below. (a) Closed-source
backbones, ReAct $\rightarrow$ TAEC with DAG agent marked, sorted by ReAct accuracy
(Table~\ref{tab:main}); the acting models are GPT-4o-mini, Gemini-3.5-Flash, Kimi-K3 and
GPT-5.6-Sol. (b) Matched subsets as a flow: the same 1{,}142 ViDoSeek
questions in the centre, grouped by the behaviour of the two systems (A: both stopped after one search;
B: ReAct kept searching while TAEC stopped; C: the remainder), with ribbons carrying each group into
``gold page found'' or ``not found'' under ReAct on the left and TAEC on the right.}
\label{fig:appendix_panels}
\end{figure}

This section holds the numbers behind the figures of the main text, so that each can be read
without the figure. Table~\ref{tab:depth_loss} is Figure~\ref{fig:evidence_loss}, the outcome
shares and the confound-free subset by trajectory depth. Table~\ref{tab:multistep_main} is
Figure~\ref{fig:results}(a), accuracy by difficulty bucket. Table~\ref{tab:matched} is the matched
subsets of Section~\ref{sec:supplementary}, Table~\ref{tab:hit_full} the coverage and use
decomposition, Table~\ref{tab:cost} the cost measurements and Table~\ref{tab:components} the
component ablation of Figure~\ref{fig:ablation}; and
Figure~\ref{fig:appendix_panels} collects two studies whose tables are named in its caption.

\begin{table}[!ht]
\centering
\caption{ReAct, unconstrained, ViDoSeek, Gemini-3.5-Flash, grouped by the number of searches the agent
issued. ``Found, unused'' is the share of questions whose gold page reached the context and whose
answer is still wrong. Right: the subset whose gold page was already among the first five retrieved
pages.}
\label{tab:depth_loss}
\vspace{3pt}
\small
\setlength{\tabcolsep}{4pt}
\begin{tabular}{lrcccc|rc}
\toprule
& \multicolumn{4}{c}{\textbf{All questions}} & & \multicolumn{2}{c}{\textbf{Gold page found at step 1}} \\
\textbf{Searches} & $n$ & Hit & Post-hit & Acc. & Found, unused & $n$ & Acc. \\
\midrule
1    & 642 & 93.8 & 96.0 & 93.6 &  3.7 & 602 & 96.0 \\
2    & 126 & 95.2 & 91.7 & 89.7 &  7.9 & 117 & 91.5 \\
3    &  71 & 80.3 & 87.7 & 77.5 &  9.9 &  37 & 83.8 \\
4--5 &  84 & 75.0 & 87.3 & 67.9 &  9.5 &  24 & 91.7 \\
6--8 &  92 & 63.0 & 63.8 & 41.3 & 22.8 &  33 & 48.5 \\
9+   & 127 & 59.1 & 52.0 & 35.4 & 28.3 &  51 & 39.2 \\
\bottomrule
\end{tabular}
\end{table}

\begin{table}[!ht]
\centering
\caption{ViDoSeek accuracy by question difficulty, Gemini-3.5-Flash. Difficulty is the number of
searches unconstrained ReAct issued on that question, so all four systems face the same questions in
each row.}
\label{tab:multistep_main}
\vspace{3pt}
\small
\setlength{\tabcolsep}{6pt}
\begin{tabular}{lrcccc}
\toprule
\textbf{Searches needed} & $n$ & \textbf{Vanilla} & \textbf{ReAct} & \textbf{DAG agent} & \textbf{TAEC} \\
\midrule
1    & 642 & 86.7 & 93.6 & 91.9 & \textbf{94.1} \\
2--3 & 197 & 78.7 & 85.3 & 79.7 & \textbf{86.3} \\
4--5 &  84 & 69.0 & 67.9 & 63.1 & \textbf{81.0} \\
6+   & 219 & 48.4 & 37.9 & 52.5 & \textbf{71.7} \\
\bottomrule
\end{tabular}
\end{table}

\begin{table}[!ht]
\centering
\caption{The questions on which ReAct kept searching, Gemini-3.5-Flash. The subset is
defined by that agent alone, so all five systems answer the same questions: 500 on ViDoSeek, 765 on
SlideVQA and 541 on MMLongBench-Doc. Coverage and the four shares are taken over the questions in
the subset that carry a reference page, which on MMLongBench-Doc is 532 of the 541. Coverage is the
sum of the first two shares. ViDoRAG records no search count, and its coverage is taken over the
pages its pipeline records for the question.}
\label{tab:matched}
\vspace{3pt}
\footnotesize
\setlength{\tabcolsep}{4.5pt}
\resizebox{\textwidth}{!}{%
\begin{tabular}{llcc|cccc}
\toprule
& & & & \multicolumn{4}{c}{\textbf{What became of the gold page (\%)}} \\
\cmidrule(lr){5-8}
\textbf{Benchmark} & \textbf{System} & \textbf{Searches} & \textbf{Coverage}
& Found, used & Found, unused & Missed, right & Missed, wrong \\
\midrule
\multirow{5}{*}{ViDoSeek}
 & Vanilla  & 1.00 & 75.0 & 58.8 & 16.2 & 5.0 & 20.0 \\
 & ReAct    & 6.00 & 74.6 & 58.2 & 16.4 & 3.4 & 22.0 \\
 & ViDoRAG  & --- & 76.0 & 61.0 & 15.0 & 4.4 & 19.6 \\
 & Pure DAG & 1.27 & 68.4 & 57.6 & 10.8 & 7.4 & 24.2 \\
 & \textbf{TAEC} & 1.28 & \textbf{90.8} & \textbf{76.6} & 14.2 & 2.4 & \textbf{6.8} \\
\midrule
\multirow{5}{*}{SlideVQA}
 & Vanilla  & 1.00 & 63.9 & 48.2 & 15.7 & 9.4 & 26.7 \\
 & ReAct    & 3.00 & 63.0 & 46.5 & 16.5 & 2.9 & 34.1 \\
 & ViDoRAG  & --- & 57.9 & 38.6 & 19.3 & 7.7 & 34.4 \\
 & Pure DAG & 1.49 & 59.3 & 42.1 & 17.3 & 9.9 & 30.7 \\
 & \textbf{TAEC} & 1.76 & \textbf{71.0} & \textbf{57.9} & 13.1 & 8.0 & \textbf{21.0} \\
\midrule
\multirow{5}{*}{MMLongBench-Doc}
 & Vanilla  & 1.00 & 32.1 & 15.8 & 16.4 & 7.5 & 60.3 \\
 & ReAct    & 3.56 & 42.3 & 21.6 & 20.7 & 2.6 & 55.1 \\
 & ViDoRAG  & --- & 34.6 & 11.1 & 23.5 & 5.1 & 60.3 \\
 & Pure DAG & 1.89 & 40.2 & 22.2 & 18.0 & 6.8 & 53.0 \\
 & \textbf{TAEC} & 2.58 & \textbf{46.2} & \textbf{28.8} & 17.5 & 5.8 & \textbf{47.9} \\
\bottomrule
\end{tabular}}
\end{table}

\begin{table}[!ht]
\centering
\caption{Accuracy decomposed into coverage and evidence use per system, ViDoSeek,
Gemini-3.5-Flash. Post-hit accuracy conditions on a quantity each system changes, so it is reported
here for completeness and is not compared across rows; the like-for-like comparison is
Figure~\ref{fig:gain_decomp}.}
\label{tab:hit_full}
\vspace{3pt}
\small
\setlength{\tabcolsep}{4pt}
\begin{tabular}{lcccc}
\toprule
\textbf{System} & \textbf{Hit rate} & \textbf{Post-hit acc.} & \textbf{Acc.\ on misses} & \textbf{Acc.} \\
\midrule
Vanilla              & 86.6 & 84.9 & 23.5 & 76.7 \\
ReAct                & 85.4 & 89.1 & 23.4 & 79.5 \\
DAG agent             & 84.4 & 90.1 & 25.8 & 80.1 \\
\textbf{TAEC}        & \textbf{94.7} & \textbf{90.6} & \textbf{31.7} & \textbf{87.5} \\
\bottomrule
\end{tabular}
\end{table}

\begin{table}[!ht]
\centering
\caption{Cost per question, ViDoSeek, Gemini-3.5-Flash, same protocol as Table~\ref{tab:main}. Cost
is what a run actually transmits: images and context are summed over every model call of a question,
so a page re-sent on a later call is counted again, and the last column is the largest single call.
Wall-clock time is not reported; it is set by concurrency and rate limits rather than by the method.}
\label{tab:cost}
\vspace{3pt}
\small
\setlength{\tabcolsep}{4pt}
\begin{tabular}{lcccc}
\toprule
\textbf{System} & \textbf{Model calls} & \textbf{Images} & \textbf{Context (MB)} & \textbf{Peak call (MB)} \\
\midrule
Vanilla       & 1.00 &  4.9 & 0.22 & 0.22 \\
ReAct         & 5.34 & 92.8 & 3.73 & 0.57 \\
DAG agent     & 4.94 & 20.0 & 0.87 & 0.23 \\
\textbf{TAEC} & \textbf{2.91} & \textbf{14.7} & \textbf{0.61} & 0.22 \\
\bottomrule
\end{tabular}
\end{table}

\begin{figure}[!ht]
\centering
\includegraphics[width=\textwidth]{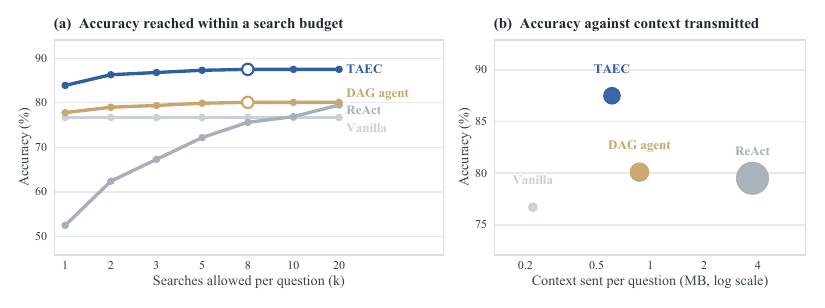}
\caption{What a trajectory spends and what the spending buys, ViDoSeek, Gemini-3.5-Flash. (a) The
accuracy a system reaches when a question is allowed at most $k$ searches, the numbers of
Table~\ref{tab:searchcap}; a ring marks the smallest budget at which a system already holds all the
accuracy it will reach. The substrate and TAEC are saturated by eight searches, while the unmanaged
agent is still gaining at twenty and reaches neither. (b) Accuracy against the context a question
transmits, the numbers of Table~\ref{tab:cost}, with point area proportional to the images sent.
TAEC answers eight points above the unmanaged agent while sending a sixth of the context and a sixth
of the images.}
\label{fig:budget_cost}
\end{figure}

\begin{table}[!ht]
\centering
\caption{Component ablation behind Figure~\ref{fig:ablation}, Gemini-3.5-Flash; the substrate and
TAEC rows are those of Table~\ref{tab:main}, and $\Delta$ is against the substrate.}
\label{tab:components}
\vspace{3pt}
\small
\setlength{\tabcolsep}{3.5pt}
\begin{tabular}{lccc|cc|cc|cc}
\toprule
& \multicolumn{3}{c|}{\textbf{Components}} & \multicolumn{2}{c|}{\textbf{ViDoSeek}} & \multicolumn{2}{c|}{\textbf{SlideVQA}} & \multicolumn{2}{c}{\textbf{MMLongBench-Doc}} \\
\textbf{Configuration} & A & M & V & Acc. & $\Delta$ & Acc. & $\Delta$ & Acc. & $\Delta$ \\
\midrule
DAG agent (substrate)    &   &   &   & 80.1 & --- & 76.7 & --- & 44.5 & --- \\
\midrule
+ Admission              & \checkmark &   &   & 86.3 & $+6.2$ & 82.3 & $+5.6$ & 48.2 & $+3.7$ \\
+ Exposure               &   & \checkmark &   & 84.1 & $+4.0$ & 82.8 & $+6.1$ & 47.2 & $+2.7$ \\
+ Allocation             &   &   & \checkmark & 83.5 & $+3.4$ & 82.4 & $+5.7$ & 46.8 & $+2.3$ \\
\midrule
\textbf{TAEC (A+M+V)}    & \checkmark & \checkmark & \checkmark & \textbf{87.5} & $+7.4$ & \textbf{83.6} & $+6.9$ & \textbf{50.1} & $+5.6$ \\
\bottomrule
\end{tabular}
\end{table}

\section{M3RAG, Reimplemented}
\label{app:m3rag}

M3RAG \citep{m3rag2026} is a published training-free method for visual document questions. A planner
decomposes the question into a graph of sub-questions, a seeker retrieves pages for each and re-reads
the region that carries the evidence, an answerer drafts a response, and a verifier accepts it or
asks for the plan to be continued, refined or rebuilt, for at most seven rounds. The authors release
no code, and four of the method's mechanisms rely on model access that the proprietary acting models
do not provide. What we run is therefore M3RAG adapted to closed backbones rather than the published
system: the planner and verifier run on the acting model rather than on a separate 7B model, the
verifier's confidence is read from a stated verdict rather than from token probabilities, region
re-reading uses a bounding box the model reports rather than patch-embedding selection, and
evidence ranking uses model-assigned relevance rather than a shared embedding space. Several
settings the paper leaves open are ours as well. We therefore report this adapted implementation
here rather than in Table~\ref{tab:main}, run against the same index, budgets and judge as every
system there.

Each of the four is replaced by the closest equivalent available.
The planner and verifier, which the paper runs on a separate
Mistral-7B, run on the acting model. The verifier's confidence, defined from token probabilities, is
read from a verdict and a confidence the model states. Region re-reading, which selects 144 visual
tokens by patch embedding, asks the model for the bounding box of the evidence and crops it at the
pixel budget of 144 tokens. Evidence ranking in a shared embedding space uses the relevance the model
assigns when it describes each item. The settings the paper states are kept: five candidate pages
per query, 144-token regions and seven rounds. For those it leaves open we use at most six
sub-questions, an evidence pool of eight, a confidence threshold of $0.5$, and replanning after two
rounds in which confidence moves by less than $0.05$.

\begin{table}[!ht]
\centering
\caption{Gemini-3.5-Flash. The other five rows are those of Table~\ref{tab:main}; \textbf{best} and
\underline{second best} per column.}
\label{tab:m3rag}
\vspace{3pt}
\small
\setlength{\tabcolsep}{6pt}
\begin{tabular}{lccc}
\toprule
\textbf{System} & \textbf{ViDoSeek} & \textbf{SlideVQA} & \textbf{MMLongBench-Doc} \\
\midrule
Vanilla              & 76.7 & 79.6 & 40.1 \\
ReAct                & 79.5 & 77.2 & 42.4 \\
ViDoRAG              & 79.7 & 72.5 & 31.5 \\
M3RAG (reimplemented) & \underline{82.5} & \underline{81.5} & \underline{47.6} \\
DAG agent            & 80.1 & 76.7 & 44.5 \\
\textbf{TAEC}        & \textbf{87.5} & \textbf{83.6} & \textbf{50.1} \\
\bottomrule
\end{tabular}
\end{table}

On Gemini-3.5-Flash M3RAG is the strongest baseline on all three benchmarks, ahead of both the
substrate and ViDoRAG. TAEC leads it by $5.0$, $2.1$ and $2.5$ points.

\begin{table}[!ht]
\centering
\caption{M3RAG against TAEC on the four acting models of Table~\ref{tab:main}, under the same index,
budgets and judge, and on the full benchmark in every cell. MMLongBench-Doc is scored on the 847
answerable questions, as in Table~\ref{tab:main}.}
\label{tab:m3rag_backbones}
\vspace{3pt}
\small
\setlength{\tabcolsep}{5pt}
\begin{tabular}{l cc c cc c cc}
\toprule
& \multicolumn{2}{c}{\textbf{ViDoSeek}} & & \multicolumn{2}{c}{\textbf{SlideVQA}}
& & \multicolumn{2}{c}{\textbf{MMLongBench-Doc}} \\
\cmidrule(lr){2-3}\cmidrule(lr){5-6}\cmidrule(lr){8-9}
\textbf{Backbone} & M3RAG & TAEC & & M3RAG & TAEC & & M3RAG & TAEC \\
\midrule
Gemini-3.5-Flash & 82.5 & \textbf{87.5} & & 81.5 & \textbf{83.6} & & 47.6 & \textbf{50.1} \\
Kimi-K3          & 80.7 & \textbf{88.1} & & 81.5 & \textbf{83.8} & & 44.5 & \textbf{45.8} \\
GPT-5.6-Sol      & 78.4 & \textbf{87.0} & & 77.3 & \textbf{82.2} & & \textbf{44.3} & 42.6 \\
GPT-4o-mini      & 49.6 & \textbf{71.0} & & 44.9 & \textbf{62.0} & & \textbf{18.7} & 17.7 \\
\bottomrule
\end{tabular}
\end{table}

Table~\ref{tab:m3rag_backbones} extends the comparison to the other three acting models and
repeats Gemini-3.5-Flash for reference. TAEC is ahead in ten of the twelve comparisons, by $1.3$ to
$21.4$ points, and behind on MMLongBench-Doc with GPT-5.6-Sol and GPT-4o-mini, by $1.7$ and $1.0$.

The two exceptions are not the same result. MMLongBench-Doc annotates $244$ of its $1{,}091$
questions as unanswerable, and Table~\ref{tab:main} scores the $847$ that carry an answer. Declining
an unanswerable question is a capability none of the three components addresses, and M3RAG's
verifier supplies it unevenly: on those $244$ it is correct on $44.3\%$ with GPT-5.6-Sol against
$16.8\%$ with Gemini-3.5-Flash, where TAEC scores $31.4\%$ and $35.8\%$. Scored over all $1{,}091$
questions (Table~\ref{tab:answerable}), the GPT-4o-mini result therefore reverses and TAEC leads by
$1.1$ points, while the GPT-5.6-Sol deficit widens to $4.2$; on the other two backbones TAEC leads
by $6.2$ and $0.9$. The GPT-5.6-Sol deficit is thus a deficit on answerable questions and not an
artefact of the question set. The main evaluation uses the benchmark's annotated answerable
subset; Appendix~\ref{app:answerable} additionally reports results on the full set.

\section{Results by Question Type}
\label{app:subtypes}

Table~\ref{tab:subtypes} breaks the main table down by the question types the benchmarks annotate.
Two observations follow from it, both on Gemini-3.5-Flash, where every cell is filled.

\paragraph{The gain is not confined to multi-hop questions.}
On ViDoSeek, TAEC improves over the substrate by $9.2$ points on single-hop questions
($77.8\rightarrow87.0$) and by $5.0$ on multi-hop ones ($83.1\rightarrow88.1$). Single-hop questions
are those a single well-chosen page answers, and they are where admitting the right page is most
consequential; multi-hop questions start from a higher base because several pages are retrieved anyway. The
same ordering holds on Kimi-K3 and GPT-4o-mini. The components therefore contribute not only the
ability to carry a long trajectory, but also the ability to spend the first retrieval well.

\paragraph{The gain is largest where the evidence is structured.}
On MMLongBench-Doc the improvement over the substrate is $+10.4$ on text evidence
($41.8\rightarrow52.2$), $+9.4$ on charts ($55.1\rightarrow64.5$) and $+4.2$ on tables
($69.2\rightarrow73.4$), but only $+3.3$ on figure evidence ($34.6\rightarrow37.9$). Figure questions
are the ones whose answer depends on reading a picture rather than on which pages are present, so
coordination has less to act on; every system is weakest there.

\begin{table}[!ht]
\centering
\caption{Accuracy by question type, same runs and judging batch as Table~\ref{tab:main}. ViDoSeek is
split by hop count; MMLongBench-Doc by the modality of the evidence, counting the questions annotated
with exactly one evidence source, so questions drawing on several modalities and the layout-only ones
enter only ``All'', which is taken over the 847 answerable questions; SlideVQA carries no question-type
annotation. Best per column within each backbone block in bold.
$\dagger$~our reimplementation, for the reasons given in Appendix~\ref{app:m3rag}.}
\label{tab:subtypes}
\vspace{3pt}
\small
\setlength{\tabcolsep}{2.6pt}
\resizebox{\textwidth}{!}{%
\begin{tabular}{llccc|c|ccccc}
\toprule
& & \multicolumn{3}{c|}{\textbf{ViDoSeek}} & \textbf{SlideVQA} & \multicolumn{5}{c}{\textbf{MMLongBench-Doc}} \\
\cmidrule(lr){3-5}\cmidrule(lr){6-6}\cmidrule(lr){7-11}
\textbf{Backbone} & \textbf{System} & Single & Multi & All & All & Text & Table & Chart & Figure & All \\
\midrule
\multirow{6}{*}{Gemini-3.5-Flash}
 & Vanilla & 71.8 & 83.1 & 76.7 & 79.6 & 41.0 & 49.7 & 55.1 & 35.2 & 40.1 \\
 & ReAct & 76.3 & 83.7 & 79.5 & 77.2 & 42.5 & 60.1 & 60.7 & 29.1 & 42.4 \\
 & ViDoRAG & 77.2 & 82.9 & 79.7 & 72.5 & 36.6 & 34.3 & 40.2 & 29.1 & 31.5 \\
 & M3RAG$^{\dagger}$ & 81.4 & 83.9 & 82.5 & 81.5 & 50.0 & 67.8 & 61.7 & 33.5 & 47.6 \\
 & DAG agent & 77.8 & 83.1 & 80.1 & 76.7 & 41.8 & 69.2 & 55.1 & 34.6 & 44.5 \\
 & \textbf{TAEC} & \textbf{87.0} & \textbf{88.1} & \textbf{87.5} & \textbf{83.6} & \textbf{52.2} & \textbf{73.4} & \textbf{64.5} & \textbf{37.9} & \textbf{50.1} \\
\midrule\multirow{6}{*}{Kimi-K3}
 & Vanilla & 81.7 & 81.9 & 81.8 & 78.0 & 34.3 & 39.2 & 44.9 & 28.0 & 31.8 \\
 & ReAct & 82.3 & 81.9 & 82.1 & 82.0 & 38.8 & 45.5 & 46.7 & 29.7 & 36.5 \\
 & ViDoRAG & 77.8 & 84.7 & 80.8 & 74.5 & 37.3 & 46.9 & 52.3 & 30.8 & 37.2 \\
 & M3RAG$^{\dagger}$ & 80.9 & 80.5 & 80.7 & 81.5 & 44.8 & \textbf{60.1} & 56.1 & \textbf{36.3} & 44.5 \\
 & DAG agent & 85.4 & 83.9 & 84.8 & 80.9 & 45.5 & 53.1 & 60.7 & \textbf{36.3} & 43.9 \\
 & \textbf{TAEC} & \textbf{89.8} & \textbf{85.9} & \textbf{88.1} & \textbf{83.8} & \textbf{47.8} & 56.6 & \textbf{61.7} & \textbf{36.3} & \textbf{45.8} \\
\midrule\multirow{6}{*}{GPT-5.6-Sol}
 & Vanilla & 80.9 & 83.5 & 82.0 & 79.3 & 39.6 & 48.3 & 46.7 & 34.6 & 38.1 \\
 & ReAct & 80.9 & 83.7 & 82.1 & 81.0 & 41.8 & 42.0 & 52.3 & 33.5 & 39.1 \\
 & ViDoRAG & 78.0 & 85.5 & 81.3 & 78.3 & 35.8 & 51.0 & \textbf{57.9} & 28.0 & 37.8 \\
 & M3RAG$^{\dagger}$ & 77.4 & 79.7 & 78.4 & 77.3 & 40.3 & \textbf{64.3} & \textbf{57.9} & 34.6 & \textbf{44.3} \\
 & DAG agent & 82.6 & \textbf{85.9} & 84.1 & 77.7 & 43.3 & 40.6 & 55.1 & 32.4 & 38.3 \\
 & \textbf{TAEC} & \textbf{87.8} & \textbf{85.9} & \textbf{87.0} & \textbf{82.2} & \textbf{47.0} & 49.0 & 54.2 & \textbf{36.8} & 42.6 \\
\midrule\multirow{6}{*}{GPT-4o-mini}
 & Vanilla & 56.4 & 62.8 & 59.2 & 61.9 & 16.4 & 14.0 & 18.7 & 19.8 & 15.6 \\
 & ReAct & 51.6 & 62.0 & 56.1 & 59.3 & 17.2 & 13.3 & 15.0 & 18.1 & 15.2 \\
 & ViDoRAG & 60.9 & \textbf{72.2} & 65.8 & 58.7 & 12.7 & 11.9 & \textbf{20.6} & 17.0 & 14.1 \\
 & M3RAG$^{\dagger}$ & 48.4 & 51.3 & 49.6 & 44.9 & \textbf{23.1} & \textbf{24.5} & 18.7 & 18.7 & \textbf{18.7} \\
 & DAG agent & 63.7 & 68.2 & 65.7 & 56.6 & 18.7 & 9.8 & 15.9 & 18.7 & 14.3 \\
 & \textbf{TAEC} & \textbf{71.6} & 70.2 & \textbf{71.0} & \textbf{62.0} & 22.4 & 12.6 & 19.6 & \textbf{20.9} & 17.7 \\
\bottomrule
\end{tabular}}
\end{table}

\subsection{MMLongBench-Doc with its unanswerable questions}
\label{app:answerable}

A quarter of MMLongBench-Doc is annotated \emph{Not answerable}: the question is posed over a
document that does not contain the answer, and a system is correct when it declines. That is a
judgement about the absence of evidence, and none of the three components acts on it; they decide
what enters the context, what stays there and how finely it is drawn. The tables of the main text
therefore score the 847 questions the benchmark annotates as answerable. The subset is the
benchmark's own label rather than a threshold of ours, and its evidence-source counts reproduce the
published ones exactly. Table~\ref{tab:answerable} gives the full set of 1{,}091 questions.

\begin{table}[!ht]
\centering
\caption{MMLongBench-Doc over all 1{,}091 questions, including the 244 annotated unanswerable, same
runs and judging batch as Table~\ref{tab:main}. Best per column in bold.
$\dagger$~our reimplementation, for the reasons given in Appendix~\ref{app:m3rag}.}
\label{tab:answerable}
\vspace{3pt}
\small
\setlength{\tabcolsep}{6pt}
\begin{tabular}{lcccc}
\toprule
\textbf{System} & \textbf{Gemini-3.5-Flash} & \textbf{Kimi-K3} & \textbf{GPT-5.6-Sol} & \textbf{GPT-4o-mini} \\
\midrule
Vanilla   & 36.8 & 32.3 & 31.5 & 22.5 \\
ReAct     & 33.3 & 32.4 & 33.5 & 16.9 \\
ViDoRAG   & 28.0 & 33.8 & 34.7 & 15.6 \\
M3RAG$^{\dagger}$ & 40.7 & 42.0 & \textbf{44.3} & 24.1 \\
DAG agent & 45.7 & 41.2 & 39.5 & 23.8 \\
\textbf{TAEC} & \textbf{46.9} & \textbf{42.9} & 40.1 & \textbf{25.2} \\
\bottomrule
\end{tabular}
\end{table}

On the full set, TAEC leads on three backbones, while M3RAG leads by $4.2$ percentage points with
GPT-5.6-Sol. The margin over the substrate is narrower here, $1.2$, $1.7$, $0.6$ and $1.4$ points,
because on an unanswerable question the correct response is to decline, which none of the
components is designed to encourage.

\section{A Trajectory in Detail}
\label{sec:case}

\begin{figure}[t]
\centering
\includegraphics[width=\textwidth]{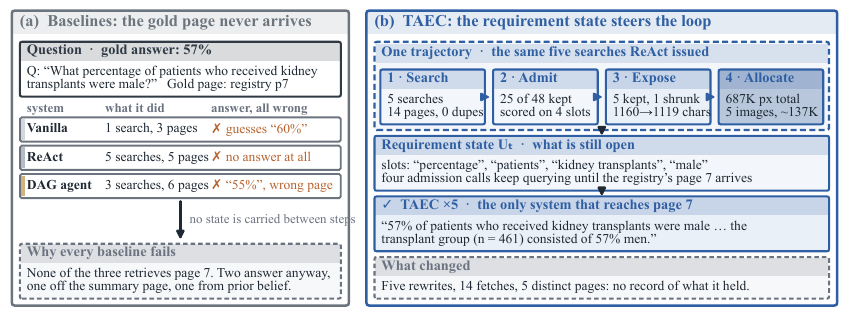}
\caption{One ViDoSeek question, Kimi-K3, from the runs behind Table~\ref{tab:main}. The answer is on
page~7 of a national renal registry report, and TAEC is the only system that retrieves it. (a) the
three baselines: none reaches the page, and two answer anyway. (b) the TAEC trajectory, the same five
searches ReAct issued: admission keeps 25 of the 48 candidates it sees across four calls scored on
requirement coverage; exposure ranks the five memories it carries by an energy that decays with their
age and compresses the one it has verified as resolved, so the carried graph is sent at 1119
characters rather than 1160; allocation spends 687K pixels over the five images exposure keeps
visible; and the requirement state carried across turns provides the basis for the next query.}
\label{fig:case}
\end{figure}

A single question illustrates the role of the requirement state inside the loop. It asks what percentage of
kidney transplant patients were male, and the answer, $57\%$, sits on page~7 of a national renal
replacement registry report. Under the same retriever and the same budget, no baseline ever retrieves
that page, and no single retrieval could have: the page is never in a top-3 result for any query any
system issues, and the first search, which is the question itself for every system but ReAct, returns
the same three pages to all of them. Vanilla states that the retrieved pages do not give the breakdown and then estimates
$60\%$ anyway; ReAct issues five searches and returns nothing; the DAG agent reaches a different page of
the same report, its summary, and reads $55\%$ off it, which is the figure for all patients rather
than for the transplant group. The pages the three trajectories retrieve separate them along the same axis on which the systems
differ. ReAct's five searches are five distinct rewrites, and they pull fourteen pages
that are only five distinct ones: holding no record of the pages it has already
retrieved, which the graph-based systems carry, its rewrites keep returning the page the retriever
ranks first for the question's own wording. The DAG agent holds that
record and repeats nothing, three searches for six distinct pages, but with no record of what is
still unresolved its queries wander inside the report and stop at the summary. TAEC holds both. Its
five searches pull fourteen distinct pages; it admits 25 of the 48 candidates it sees across four
admission calls scored on the four open requirements, admitting candidates whose requirement coverage averages
$0.35$ against $0.25$ in the pool they are drawn from; it carries five memories under an age-decayed
exposure that compresses the one it has verified as resolved, so that five of the fourteen pages are
still visible as images when the answer is written, and spends 687K pixels over those five. Page~7 arrives eleventh, on a query written from the slots that are
still open, and the answer cites the transplant group of 461 patients. The search count is matched
with ReAct, so what separates them is not how much they search but what each search is written
against. This is the coverage group of Figure~\ref{fig:results}(b) at the
scale of a single question.

\section{Context Pressure with Retrieval Held Fixed}
\label{app:interference}

Section~\ref{sec:diagnosis} reads accuracy against the depth a trajectory reaches, and depth and
difficulty move together there. This study separates them. Retrieval is run once and then frozen:
every configuration answers the same questions from the same retrieved pages, with the annotated
evidence page present in the context throughout. The only quantity that varies is the amount of
neutral text placed alongside it, drawn from documents unrelated to the question and carrying no
evidence bearing on it.

\begin{table}[H]
\centering
\caption{The substrate on ViDoSeek, Gemini-3.5-Flash, with retrieval frozen and the gold page in the
context in every column.}
\label{tab:interference}
\vspace{3pt}
\small
\setlength{\tabcolsep}{8pt}
\begin{tabular}{lccc}
\toprule
\textbf{Neutral tokens added} & 0 & 16K & 32K \\
\midrule
Accuracy & 78.8 & 74.8 & 73.6 \\
\bottomrule
\end{tabular}
\end{table}

Accuracy falls by $4.0$ points over the first 16K tokens and by $5.2$ over 32K. Nothing about the
evidence changes across the row: the same pages are in the context, rendered the same way, and the
question is the same. What changes is how much else the model reads before it reaches them.

This is the half of the diagnosis that the depth reading cannot supply. A system that searches more
also retrieves more, so an accuracy decline with depth admits a retrieval explanation; here retrieval
is held constant by construction and the decline survives. The evidence a question needs can be
present and still be read less reliably, which is the condition the three components are introduced
to manage.

\section{Where the Multi-Step Baselines Lose}
\label{app:depth}

Table~\ref{tab:main} has four columns in which ReAct scores below single-pass retrieval.
Table~\ref{tab:reactdepth} locates the loss. Each benchmark is split by the number of
searches ReAct issued on that question, an index of its own behaviour, and single-pass retrieval is
scored on the same two question sets, so the two systems always answer the same questions.

\begin{table}[!ht]
\centering
\caption{ReAct against single-pass retrieval on the same questions, split by the depth ReAct chose.
$\Delta$ is ReAct minus Vanilla. The rows are the four cells of Table~\ref{tab:main} where ReAct
falls below single-pass retrieval, together with three where it does not: Gemini-3.5-Flash on
ViDoSeek and MMLongBench-Doc, and Kimi-K3 on SlideVQA.}
\label{tab:reactdepth}
\vspace{3pt}
\small
\setlength{\tabcolsep}{3.5pt}
\begin{tabular}{llcccc|cccc}
\toprule
& & \multicolumn{4}{c|}{\textbf{ReAct stopped after one search}} & \multicolumn{4}{c}{\textbf{ReAct searched again}} \\
\cmidrule(lr){3-6}\cmidrule(lr){7-10}
\textbf{Benchmark} & \textbf{Backbone} & $n$ & ReAct & Vanilla & $\Delta$ & $n$ & ReAct & Vanilla & $\Delta$ \\
\midrule
\multirow{2}{*}{ViDoSeek}
 & Gemini-3.5-Flash &  642 & 93.5 & 86.8 & $+6.7$ &  500 & 61.6 & 63.8 & $-2.2$ \\
 & GPT-4o-mini      &  910 & 62.2 & 63.8 & $-1.6$ &  232 & 32.3 & 40.9 & $-8.6$ \\
\midrule
\multirow{3}{*}{SlideVQA}
 & Gemini-3.5-Flash & 1450 & 91.8 & 91.2 & $+0.6$ &  765 & 49.4 & 57.6 & $-8.2$ \\
 & Kimi-K3          & 1762 & 88.6 & 86.2 & $+2.4$ &  453 & 56.3 & 46.1 & $+10.2$ \\
 & GPT-4o-mini      & 1827 & 62.6 & 65.7 & $-3.1$ &  388 & 44.1 & 44.3 & $-0.3$ \\
\midrule
\multirow{2}{*}{MMLongBench-Doc}
 & Gemini-3.5-Flash &  306 & 74.8 & 69.9 & $+4.9$ &  541 & 24.0 & 23.3 & $+0.7$ \\
 & GPT-4o-mini      &  583 & 18.2 & 19.4 & $-1.2$ &  264 &  8.7 &  7.2 & $+1.5$ \\
\bottomrule
\end{tabular}
\end{table}

Two readings follow. First, the large deficits are a property of the deeper half. Where ReAct answered after
one search it is within a few points of single-pass retrieval and often above it, which is expected,
since on those questions the two systems do nearly the same thing. Where it searched again it loses
$8.6$ points on ViDoSeek with GPT-4o-mini and $8.2$ on SlideVQA with Gemini-3.5-Flash, so the additional searches neither recovered evidence
the first one missed nor left the rest of the question untouched. The cost measurements of
Table~\ref{tab:cost} give the mechanism a size: ReAct re-sends its observation history on every call
and reaches 93 images per question, and accuracy falls with the number of images in context even
when the gold page is already among them.

Second, the sign is not a property of the method alone. On Kimi-K3 the same loop gains $10.2$ points
on the deeper half of SlideVQA. What separates the rows is whether the acting model both issues
useful follow-up searches and still reads what accumulates; GPT-4o-mini fails the second condition on
ViDoSeek and SlideVQA, and the open-weight backbones of Appendix~\ref{app:open} fail the first as
well. An unmanaged trajectory therefore leaves the outcome to the acting model, which is the
variance the substrate and the components are introduced to remove.

\section{How the Acting Model Determines the Size of the Gain}
\label{app:frontier}

Two abilities of an acting model govern how much the components add to it, and the substrate
separates them. Every substrate row runs one body of code, one retriever and one budget, so a
difference between its columns is an ability of the model alone.

The first is perceptual: converting a page placed in front of it into an answer. It is measured as
accuracy on the questions the model answered after a single search whose gold page was in the
context. The second is agentic: judging, at each step of a trajectory, whether what has been
accumulated answers the question \citep{flare2023,selfrag2024}. That judgement is what a multi-step
system asks of the acting model at every turn, and the trajectory it produces is where the judgement
becomes observable. It fails in two directions. A model commits to an answer on evidence that does
not support it, which appears as a trajectory ended with no gold page in the context. A model fails
to recognise evidence that already answers the question, which appears as a further search on a
question a single retrieval had answered. Table~\ref{tab:backbonegain} reports the perceptual ability
and both directions of the agentic one, together with the accuracy gaps of Table~\ref{tab:main}, each
averaged over the three benchmarks.

\begin{table}[!ht]
\centering
\caption{Each column describes one acting model, averaged over the three benchmarks of
Table~\ref{tab:main}. The first block is measured on the substrate, where the four columns differ in
nothing but the model. \emph{Reads a retrieved gold page} is accuracy on the single-search questions
whose gold page reached the context. The two agentic rows are the two directions in which a
sufficiency judgement fails: \emph{commits without a gold page} is the share of all questions
answered after one search with no gold page in the context, and \emph{over-searches an answered
question} is the share of the questions whose gold page a single retrieval already reached on which
the system issued a further search.}
\label{tab:backbonegain}
\vspace{3pt}
\small
\setlength{\tabcolsep}{5pt}
\begin{tabular}{lcccc}
\toprule
& \textbf{Gemini-3.5-Flash} & \textbf{Kimi-K3} & \textbf{GPT-5.6-Sol} & \textbf{GPT-4o-mini} \\
\midrule
\multicolumn{5}{l}{\emph{Perceptual ability, on the substrate}}\\
Reads a retrieved gold page        & 85.0 & 85.3 & 82.5 & 64.2 \\
\multicolumn{5}{l}{\emph{Agentic ability, on the substrate}}\\
Commits without a gold page        & 22.6 & 12.2 & 14.8 & 9.5 \\
Over-searches an answered question & 6.0 & 15.5 & 12.8 & 47.9 \\
Searches per question              & 1.33 & 1.72 & 1.62 & 3.10 \\
\midrule
\multicolumn{5}{l}{\emph{Agentic ability, under coordination}}\\
Commits without a gold page        & 9.2 & 9.3 & 8.2 & 6.5 \\
Over-searches an answered question & 6.0 & 6.1 & 6.6 & 40.0 \\
Searches per question              & 1.54 & 1.72 & 1.84 & 2.42 \\
\midrule
\multicolumn{5}{l}{\emph{Accuracy against the row below on the trajectory axis}}\\
Single-pass accuracy         & 65.5 & 63.9 & 66.5 & 45.6 \\
ReAct $-$ Vanilla            & $+0.9$ & $+3.0$ & $+0.9$ & $-2.0$ \\
DAG agent $-$ Vanilla        & $+1.6$ & $+6.0$ & $+0.2$ & $0.0$ \\
TAEC $-$ Vanilla             & $+8.3$ & $\mathbf{+8.7}$ & $+4.1$ & $+4.7$ \\
TAEC $-$ DAG agent           & $\mathbf{+6.6}$ & $+2.7$ & $+3.9$ & $+4.7$ \\
\bottomrule
\end{tabular}
\end{table}

The three strongest backbones read at one level and judge sufficiency at three different ones.
Perceptual accuracy is $85.0$, $85.3$ and $82.5$, so what separates their columns is the agentic
ability alone, and each errs in its own direction. Gemini-3.5-Flash commits early on $22.6\%$ of
questions and over-searches an answerable one on $6.0\%$; Kimi-K3 reverses both figures, at $12.2$
and $15.5$; GPT-5.6-Sol lies between them at $14.8$ and $12.8$.

Coordination narrows both directions to a common level: premature commitment falls to $9.2$, $9.3$
and $8.2$, and unrecognised sufficiency to $6.0$, $6.1$ and $6.6$. Two quantities that span $10.4$ and
$9.5$ points across these backbones close to about one point each. The components leave the judgement itself
with the acting model and change the evidence it is made on, which is why the accuracy they add
orders as the size of the repair rather than as the capability of the model: $+6.6$ on
Gemini-3.5-Flash, whose larger error falls furthest, and $+3.9$ and $+2.7$ on GPT-5.6-Sol and
Kimi-K3, which begin closer to that level.

GPT-4o-mini is limited in both abilities at once, and shows what each limit costs. It reads a
retrieved gold page twenty points below the others, at $64.2$, which bounds what any improvement in
presentation can buy. Its sufficiency judgement is also the one that responds least to the evidence
it is made on: it searches again on $47.9\%$ of the questions a single retrieval had already
answered, three times the rate of any backbone above, and coordination moves that to $40.0$ against
falls to near $6$ elsewhere. The components still add $4.7$ points to it, drawn from the admission of
evidence rather than from the course of the trajectory.

Coordination also lengthens the trajectory, on Gemini-3.5-Flash from $1.33$ searches to $1.54$, and
the evidence gain holds at a matched length. Among the questions each system answered after a single
search, a gold page is in the context for $96.4\%$ of them under coordination against $85.9\%$ on the
substrate on ViDoSeek, and $92.7\%$ against $80.4\%$ on SlideVQA, with no more pages in context in
either case.

A model that reads well and whose sufficiency judgement errs towards committing early is the case a
coordination layer is for, and Gemini-3.5-Flash is the clearest instance of it among the four. The
open-weight backbone of Appendix~\ref{app:open} extends the direction GPT-4o-mini already marks. On
Qwen2.5-VL-7B that judgement is close to detached from the evidence: the unmanaged loop costs $28.8$
points on ViDoSeek and $23.2$ on SlideVQA, it answers $24.2\%$ of ViDoSeek questions without
searching at all, and under coordination it continues searching on $58.1\%$ of questions where
Gemini-3.5-Flash continues on $6.2\%$. The components add $2.4$ points to the substrate there, and
single-pass retrieval remains the strongest configuration for every multi-step system on it.

The two abilities bound the gain from two sides. Perception sets what an improved presentation of
evidence can be worth, and the sufficiency judgement sets how much of that worth a trajectory keeps.
These components govern the evidence on which that judgement is made, and the judgement stays with
the acting model.

Three further patterns hold across the cells themselves. Among the methods of
Table~\ref{tab:main}, TAEC is ahead in every cell. The margin over single-pass retrieval is largest on ViDoSeek and MMLongBench-Doc, the two
benchmarks whose questions most often need more than one search, and smallest on SlideVQA, whose questions are usually answered from one slide. The spread of the
ViDoRAG row across backbones is of the same order, which is consistent with a multi-agent workflow
that also carries its retrieved pages forward.

\section{Open-Weight Backbones}
\label{app:open}

Every number in the main text uses a proprietary API backbone. This appendix repeats the protocol on
two open-weight ones, both to report results that others can reproduce without an API and to give the
limitation stated in the conclusion its evidence. The finding is that the components require an
acting model able to run its own trajectory, and that this is a different property from reading a
page well.

\paragraph{Deployment.}
Qwen2.5-VL-7B-Instruct is served locally with vLLM on two NVIDIA A100-80GB GPUs, tensor parallel 2,
native tool calling, at most ten images per prompt; the retrieval service runs on the same machine.
Every other setting, including the retriever, the judge, the step and visual budgets and the prompts,
is identical to the main protocol (Section~\ref{sec:setup}). The RL-trained VRAG-RL checkpoint and
Qwen3-VL-4B-Instruct are served the same way.

\subsection{Results on open-weight backbones}
\label{app:backbones}

\begin{table}[!ht]
\centering
\caption{Accuracy on open-weight backbones, all rows judged with the prompt of Table~\ref{tab:main}
and run under the same harness; as in Table~\ref{tab:main}, MMLongBench-Doc is scored on its 847
answerable questions. ViDoRAG was run on Qwen3-VL-4B only. The VRAG-RL row uses the released checkpoint as the acting model in our harness;
its own loop does not transfer to our tool interface, which accounts for the value in the ReAct
column. $\dagger$~our reimplementation, for the reasons given in Appendix~\ref{app:m3rag}; its
planner and verifier run on the acting model, which is what the Qwen2.5-VL-7B row reflects.}
\label{tab:open}
\vspace{3pt}
\small
\setlength{\tabcolsep}{5pt}
\resizebox{\textwidth}{!}{%
\begin{tabular}{llcccccc}
\toprule
\textbf{Backbone} & \textbf{Benchmark} & \textbf{Vanilla} & \textbf{ReAct} & \textbf{ViDoRAG} & \textbf{M3RAG}$^{\dagger}$ & \textbf{DAG agent} & \textbf{TAEC} \\
\midrule
\multirow{2}{*}{Qwen2.5-VL-7B}
 & ViDoSeek & \textbf{67.9} & 39.1 & --- & 22.1 & 63.9 & 66.3 \\
 & SlideVQA & \textbf{57.0} & 33.8 & --- & --- & 55.3 & 55.4 \\
\midrule
\multirow{3}{*}{Qwen3-VL-4B}
 & ViDoSeek & 68.1 & 63.4 & 70.2 & 58.1 & 68.0 & \textbf{71.5} \\
 & SlideVQA & 58.3 & 59.0 & \textbf{62.8} & 51.5 & 60.5 & 60.4 \\
 & MMLongBench-Doc & 15.0 & 15.8 & 4.7 & 14.5 & 14.8 & \textbf{16.1} \\
\midrule
VRAG-RL 7B (RL) & ViDoSeek & --- & 12.7 & --- & --- & 58.0 & \textbf{58.7} \\
\bottomrule
\end{tabular}}
\end{table}

Two readings, and they point in opposite directions. The components do add to the substrate here, by
$2.4$ points on ViDoSeek and $0.7$ on the RL-tuned checkpoint, so the operators themselves are not
inert on a small model. But single-pass retrieval is the strongest configuration on both benchmarks,
and the reason is visible in how the trajectories run rather than in how the pages are read. As
ReAct, this backbone answers $24.2\%$ of ViDoSeek questions without issuing a retrieval at all,
against under $1\%$ for every proprietary backbone in Table~\ref{tab:main}; under TAEC it continues
searching on $58.1\%$ of questions, where Gemini-3.5-Flash continues on $6.2\%$.

The newer and smaller Qwen3-VL-4B behaves differently. TAEC is the strongest configuration on
ViDoSeek and MMLongBench-Doc; on SlideVQA ViDoRAG leads, and TAEC is one question behind the
substrate. The model seldom continues
searching: under TAEC it answers $97.4\%$ of ViDoSeek questions after one search, $84.8\%$ of SlideVQA
and $62.2\%$ of MMLongBench-Doc. What the components change on this backbone is therefore mainly which
pages the first search admits, and the gain is largest on the benchmark where that search decides
the most, ViDoSeek.

Table~\ref{tab:stopcost} prices that. It keeps only the questions a single search already answered
in the sense that matters, those whose annotated page single-pass retrieval retrieved, and groups
them by how many searches TAEC then issued. Single-pass accuracy is almost flat down the table, so
the groups are of comparable difficulty; ours falls with every additional search.

\begin{table}[!ht]
\centering
\caption{Qwen2.5-VL-7B on the 989 ViDoSeek questions whose annotated page single-pass retrieval
reached, grouped by the number of searches TAEC issued on them. The evidence was present after one
search in every row.}
\label{tab:stopcost}
\vspace{3pt}
\small
\setlength{\tabcolsep}{6pt}
\begin{tabular}{lcccc}
\toprule
\textbf{Searches TAEC issued} & $n$ & \textbf{Vanilla} & \textbf{TAEC} & $\Delta$ \\
\midrule
1        & 442 & 80.3 & 78.7 & $-1.6$ \\
2        & 139 & 76.3 & 71.2 & $-5.1$ \\
3--4     & 178 & 76.4 & 65.7 & $\mathbf{-10.7}$ \\
5+       & 230 & 69.6 & 62.2 & $-7.4$ \\
\bottomrule
\end{tabular}
\end{table}

The evidence was in the context after the first search in all four rows, and the trajectory kept
going. That is a decision the components do not make: admission, exposure and allocation shape what
the acting model is shown, and stopping is left to the model. On a backbone that stops when it
should, the same three components are worth $6.1$ points over single-pass retrieval on the
corresponding subset; on this one the trajectory spends what they win.

\subsection{Trajectory depth and component footprint on an open-weight backbone}
\label{app:weak}

The room to recover is larger here and the recovery does not follow: averaged over the benchmarks
the components add $0.4$ points against $2.7$ to $6.6$ on the closed-source backbones of
Table~\ref{tab:backbonegain}, because a model that answers a quarter of questions without searching
is not in a position to use a better-managed trajectory.

\subsection{Comparison with RL-trained systems}
\label{app:rl}

TAEC is training-free and is compared in the main text only against training-free systems. The
recent RL-trained visual RAG agents, VRAG-RL \citep{vragrl2025}, VISOR \citep{visor2026}
and VimRAG \citep{vimrag2026}, train a 3B--8B policy to search, crop and answer, and report on
Qwen2.5-VL-7B. They differ from TAEC in what is being improved: RL changes the policy that decides
what to search and when to stop; TAEC changes what the policy is shown, how its memory is exposed and
where its pixels go, and leaves the policy fixed. The two therefore act on the two abilities
Appendix~\ref{app:frontier} separates, steering a trajectory and reading what accumulates, and this
backbone is the one whose first ability fails. Table~\ref{tab:rl} places both on it.

\begin{table}[!ht]
\centering
\caption{Qwen2.5-VL-7B on ViDoSeek. Published rows are taken from the VISOR table and use its
judge, retriever and page collection; ours use the protocol of Table~\ref{tab:main}. The two blocks
are indicative of what each line of work reports on this backbone and are not comparable cell by
cell. The bottom block uses the released VRAG-RL checkpoint
as the acting model in our harness; its own ReAct-style loop does not transfer to our tool interface,
which accounts for the low value in the first row.}
\label{tab:rl}
\vspace{3pt}
\small
\setlength{\tabcolsep}{4pt}
\begin{tabular}{llcc}
\toprule
\textbf{System} & \textbf{Training} & \textbf{ViDoSeek} & \textbf{Source} \\
\midrule
Vanilla                & none & 32.9 & \citet{visor2026} \\
ReAct                  & none & 33.9 & \citet{visor2026} \\
ReAct                  & none & 39.1 & ours \\
ViDoRAG                & none & 69.0 & \citet{visor2026} \\
M3RAG                  & none & 69.4 & \citet{visor2026} \\
\textbf{TAEC}          & none & 66.3 & ours \\
\midrule
VRAG-RL                & RL   & 66.8 & \citet{visor2026} \\
EVisRAG                & RL   & 69.8 & \citet{visor2026} \\
VISOR                  & RL   & \textbf{74.9} & \citet{visor2026} \\
\midrule
VRAG-RL checkpoint as ReAct       & RL         & 12.7 & ours \\
VRAG-RL checkpoint on substrate   & RL         & 58.0 & ours \\
VRAG-RL checkpoint + TAEC         & RL + TAEC  & 58.7 & ours \\
\bottomrule
\end{tabular}
\end{table}

Two observations. First, the RL systems' gains come from a trained search policy, and against the
strongest training-free system in the table they range from $2.6$ points behind to $5.5$ ahead, at a
training cost TAEC does not pay.
Second, composition is possible here but small: on the released checkpoint the coordination layer
adds $0.7$ points over the same checkpoint on the bare substrate. The reading that fits the rest of
this appendix is that training the search policy addresses one of the two abilities a coordination
layer needs from its acting model and leaves the other where it was. The checkpoint is a 7B model
throughout, and what it reads off a page does not change when its policy is trained; a layer that
improves how evidence is presented can only be worth what the model can then read. We report the
figure as the composition we measured on the one public checkpoint available, not as the size of the
effect in general.

Third, what the table does not establish. The two blocks are produced by different judges,
retrievers and page collections, so no row in one is an estimate of what the other would score under
its conditions, and we make no claim that coordination is worth less than a trained search policy.
The backbone the RL literature reports on is the one backbone in this paper whose trajectory control
fails, and on it the quantity a coordination layer improves has little left to act through: by
Table~\ref{tab:stopcost} the trajectory has already spent what the components win before the answer
is generated. A comparison drawn here would measure the value of training that control, which is not
in question, rather than the value of coordinating evidence, which is what the rest of the paper
measures on backbones where control holds. The two approaches are addressed to different halves of
the same problem, and this backbone is not the setting in which to weigh one against the other.

\section{Retriever Control}
\label{app:retrievers}

Every table in the paper holds the retriever fixed, which leaves open whether the components depend
on the one we chose. Table~\ref{tab:retrievers} repeats three systems on retrievers of different
strength and kind: the dense index of the main protocol, a BM25 index over page text, and a hybrid
that adds link structure to the dense scores.

\begin{table}[!ht]
\centering
\caption{The same three systems on retrievers of different strength, ViDoSeek, Gemini-3.5-Flash, same
judging batch as the main tables. The last column is TAEC minus the stronger baseline in the row.}
\label{tab:retrievers}
\vspace{3pt}
\small
\setlength{\tabcolsep}{4pt}
\begin{tabular}{lcccc}
\toprule
\textbf{Retriever} & \textbf{Vanilla} & \textbf{DAG agent} & \textbf{TAEC} & \textbf{$\Delta$} \\
\midrule
Dense (main)                      & 76.7 & 80.1 & \textbf{87.5} & $+7.4$ \\
BM25 over page text               & 72.8 & 80.2 & \textbf{80.7} & $+0.5$ \\
Hybrid (dense + link structure)   & 82.7 & 78.5 & \textbf{87.1} & $+4.4$ \\
\bottomrule
\end{tabular}
\end{table}

TAEC leads on all three, and the ordering of the baselines changes with the retriever while the
ordering of TAEC against them does not. The size of the lead does change, and it tracks how much
the retriever leaves for selection to do. The dense index of the main protocol leaves the most, and
the lead is $7.4$ points. BM25 leaves the least: it matches page captions lexically, so it cannot
reach the semantic variants that requirement-driven rewriting produces, and the density of gold
pages in its candidate pool is low enough that choosing among them for coverage has little to
amplify, which leaves $0.5$. On the hybrid index single-pass retrieval is itself $6.0$ points
stronger than on the dense one, so less of the question is left above what one query already
returns; both trajectory systems sit slightly below their dense values there and TAEC keeps a $4.4$
point lead. What transfers across the three is that coordination is worth having; what depends on
the retriever is how much of a pool it has to work with.

\section{Search Depth}
\label{app:searchcap}

The main protocol allows at most 20 model calls per question. Table~\ref{tab:searchcap} reads the
recorded trajectories as a cumulative curve: for
each $k$ it reports the accuracy a system has already attained using at most $k$ searches, so the
rows show how early each system realises its final accuracy rather than how it would behave under an
imposed limit.

At small $k$ a value combines two things, how accurately a system reads and how early it decides to
stop, and the questions it has stopped on are the ones it judged answerable after a single search.
Vanilla is the only column free of this, since it always issues exactly one search and its value is
its full accuracy. The controlled comparison of what depth contributes is in
Section~\ref{sec:supplementary}, where questions are stratified by an external measure of difficulty
so that every system faces the same questions in each bucket; the table here is about saturation.

\begin{table}[!ht]
\centering
\caption{Accuracy attained within $k$ searches, ViDoSeek, Gemini-3.5-Flash, same runs and judging
batch as Table~\ref{tab:main}. A question counts when the system answered it using at most $k$
searches and answered it correctly. The table reads the multi-step range, from two searches upward,
where the components are active. No TAEC trajectory exceeds eight searches and no substrate
trajectory exceeds seven, so both are saturated from eight on, while ReAct continues to gain up to
nineteen and is still two and a half points short at ten.}
\label{tab:searchcap}
\vspace{3pt}
\small
\setlength{\tabcolsep}{4pt}
\begin{tabular}{lcccc}
\toprule
\textbf{Searches used $\le k$} & \textbf{Vanilla} & \textbf{ReAct} & \textbf{DAG agent} & \textbf{TAEC} \\
\midrule
2  & 76.7 & 62.4 & 79.0 & \textbf{86.3} \\
3  & 76.7 & 67.3 & 79.4 & \textbf{86.8} \\
5  & 76.7 & 72.2 & 79.9 & \textbf{87.3} \\
8  & 76.7 & 75.6 & 80.1 & \textbf{87.5} \\
10 & 76.7 & 76.9 & 80.1 & \textbf{87.5} \\
20 & 76.7 & 79.5 & 80.1 & \textbf{87.5} \\
\bottomrule
\end{tabular}
\end{table}

\section{Why Post-hit Accuracy Cannot Be Compared Across Systems}
\label{app:gaindecomp}

Post-hit accuracy, the accuracy on the questions for which a system retrieved a gold page, is the
natural way to ask whether a system uses the evidence it finds. It cannot be compared between two
systems, because its denominator is exactly what the method changes: the system that retrieves more
admits the harder questions into its own denominator and is penalised for doing so. On ViDoSeek the substrate scores $90.1$ on its
own hits and the full system $90.6$ on its own, a gap of less than a point, while their accuracies
differ by $7.4$; the quantity is almost uninformative about the mechanism because the two denominators
are different sets of questions.

\begin{figure}[!ht]
\centering
\includegraphics[width=\textwidth]{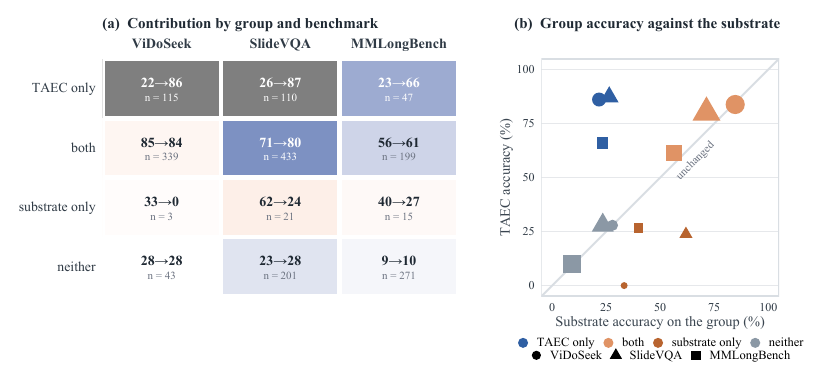}
\caption{The four-group decomposition, all three benchmarks, Gemini-3.5-Flash. (a) One cell per
group and benchmark: the substrate's accuracy and TAEC's on that group, and its size; fill is the
group's contribution to the accuracy difference. (b) The same twelve groups against the identity
line, point area $\propto$ group size.}
\label{fig:gain_decomp}
\end{figure}

The fix is to group questions by what both systems retrieved and compare inside a group, where the
two systems answer identical questions. The four groups are disjoint, and each one's difference
weighted by its size sums exactly to the difference in accuracy, so the decomposition is an identity
rather than an approximation.
Figure~\ref{fig:gain_decomp} gives all four groups on all three benchmarks, computed from the
per-question records of the runs behind Table~\ref{tab:main}; Figure~\ref{fig:results}(b) carries the
two that the gain comes from.

Two things follow that the main-text view cannot show. Coverage is close to nested: the two systems
retrieve differently on 118, 131 and 62 questions, of which 115, 110 and 47 are the ones only TAEC
retrieves, so coordination almost never loses a page the substrate found, and the reverse groups are
small enough to be noise, three questions on ViDoSeek. And a retrieval miss is not an automatic loss:
with no gold page in context the substrate still answers $28$, $23$ and $9$ percent of those
questions correctly and TAEC $28$, $28$ and $10$, which is the floor any coverage number has to be
read against.

\section{Protocol Differences Behind the Baseline Gap}
\label{app:audit}

Three differences between this protocol and those of the papers we compare against affect how the
numbers should be read. All of them apply to every row of a table alike, so comparisons within a
table remain like-for-like; across papers they mean our values are not directly comparable to
published ones.

\begin{table}[!ht]
\centering
\caption{Protocol differences between this paper and the works we compare against, and how each
affects the numbers. Rows from different retriever fingerprints are never placed in one table.}
\label{tab:audit}
\vspace{3pt}
\small
\setlength{\tabcolsep}{5pt}
\begin{tabular}{p{0.30\textwidth}p{0.50\textwidth}}
\toprule
\textbf{Difference} & \textbf{Effect on the reported numbers} \\
\midrule
Retrieval index and pages placed per search held common to every system & Published numbers come from per-system
indices of different strength; ours are read against one pooled index of 36,233 pages \\
One judge and one prompt for every row & ViDoRAG grades on a five-point scale with GPT-4o and counts
four or above; an in-run 7B judge scores the same predictions 37.9 where GPT-4.1 scores 80.1 \\
Answers scored as returned & A trajectory that terminates without an answer is scored on a
plain-text fallback rather than as zero, which is worth $8.5$ points to ReAct \\
\bottomrule
\end{tabular}
\end{table}

We list these because a reader comparing against published numbers on these benchmarks will meet the
same pitfalls, and because three of them were ours. The retriever finding is the most consequential:
the search URL is not an identifier of the retriever, and without a behavioural fingerprint two runs a
week apart can differ by more than any method effect in this paper.

\end{document}